\documentclass[preprint,sort&compress, 12pt]{elsarticle}
\usepackage{lipsum}
\makeatletter

\def\ps@pprintTitle{%
 \let\@oddhead\@empty
 \let\@evenhead\@empty
 \def\@oddfoot{}%
 \let\@evenfoot\@oddfoot}
\makeatother

\usepackage[utf8]{inputenc}
\usepackage{algpseudocode}
\usepackage{algorithm}
\usepackage[margin=1in]{geometry}
\usepackage{bold-extra}
\usepackage{xcolor}
\usepackage{adjustbox}
\usepackage{appendix}
\usepackage{lineno}

\usepackage{amssymb}
\usepackage{amsthm}
\usepackage{amsmath}
\usepackage{graphicx}
\usepackage{verbatim}
\usepackage{times}
\usepackage[titles]{tocloft}
\usepackage{enumerate}
\usepackage{bm}
\usepackage{sidecap}
\usepackage{epstopdf}
\usepackage{pdfpages}
\usepackage[colorinlistoftodos]{todonotes}
\usepackage{soul}
\usepackage{multicol}
\usepackage{float}
\usepackage{multibib}
\usepackage{geometry}
\usepackage[small,compact]{titlesec}
\usepackage[normal,bf,labelsep=colon,skip=2pt]{caption}
\usepackage{setspace}
\usepackage{subfig}
\usepackage{tocloft}
\usepackage{url}
\usepackage{floatflt}
\usepackage{array}
\usepackage{pgfgantt}
\usepackage{sidecap}
\usepackage{cite}
\usepackage{hyperref}
\usepackage{multirow}

\numberwithin{equation}{section}
\numberwithin{figure}{section}
\numberwithin{table}{section}

\graphicspath{{./plots/}}

\newcommand*\diff{\mathop{}\!\mathrm{d}}

\def\d{{\, \rm d}}
\def\d{{\, \rm d}}

\begin{document}

\begin{frontmatter}

\title{CGNSDE: Conditional Gaussian Neural Stochastic Differential Equation for Modeling Complex Systems and Data Assimilation}

\author[1]{Chuanqi Chen}
\ead{cchen656@wisc.edu}
\author[2]{Nan Chen}
\ead{chennan@math.wisc.edu}
\author[1]{Jin-Long Wu}
\ead{jinlong.wu@wisc.edu}

\address[1]{Department of Mechanical Engineering, University of Wisconsin–Madison, Madison, WI 53706}
\address[2]{Department of Mathematics, University of Wisconsin–Madison, Madison, WI 53706}

\begin{abstract}
A new knowledge-based and machine learning hybrid modeling approach, called conditional Gaussian neural stochastic differential equation (CGNSDE), is developed to facilitate modeling complex dynamical systems and implementing analytic formulae of the associated data assimilation (DA). In contrast to the standard neural network predictive models, the CGNSDE is designed to effectively tackle both forward prediction tasks and inverse state estimation problems. The CGNSDE starts by exploiting a systematic causal inference via information theory to build a simple knowledge-based nonlinear model that nevertheless captures as much explainable physics as possible. Then, neural networks are supplemented to the knowledge-based model in a specific way, which not only characterizes the remaining features that are challenging to model with simple forms but also advances the use of analytic formulae to efficiently compute the nonlinear DA solution. These analytic formulae are used as an additional computationally affordable loss to train the neural networks that directly improve the DA accuracy. This DA loss function promotes the CGNSDE to capture the interactions between state variables and thus advances its modeling skills. With the DA loss, the CGNSDE is more capable of estimating extreme events and quantifying the associated uncertainty. Furthermore, crucial physical properties in many complex systems, such as the translate-invariant local dependence of state variables, can significantly simplify the neural network structures and facilitate the CGNSDE to be applied to high-dimensional systems. Numerical experiments based on chaotic systems with intermittency and strong non-Gaussian features indicate that the CGNSDE outperforms knowledge-based regression models, and the DA loss further enhances the modeling skills of the CGNSDE.
\end{abstract}

\begin{keyword}
Complex dynamical systems \sep Machine learning \sep Data assimilation \sep Uncertainty quantification \sep Causal inference \sep Analytically solvable statistics
\end{keyword}

\end{frontmatter}

\section{Introduction}

Complex dynamical systems appear in many areas, including geophysics, climate science, engineering, neural science, and material science \citep{jost2005dynamical, frisch1995turbulence, majda2006nonlinear, vallis2017atmospheric, abraham1984complex, salmon1998lectures, brin2002introduction, strogatz2018nonlinear, wiggins2003introduction, chen2023stochastic}. These systems are highly nonlinear and are often strongly chaotic or turbulent \citep{salmon1998lectures, dijkstra2013nonlinear, palmer1993nonlinear}. Intermittency, extreme events, and non-Gaussian probability density functions (PDFs) are some of the typical features in these systems \citep{farazmand2019extreme, trenberth2015attribution, moffatt2021extreme, majda2003introduction, manneville1979intermittency}. Developing appropriate models to describe the observed features of these complex systems is crucial in exploring the underlying physics and simulating the associated phenomena. These models also play an essential role in estimating the states via data assimilation (DA) \citep{kalnay2003atmospheric, lahoz2010data, majda2012filtering, evensen2009data, law2015data}, which is the pre-requite for statistical forecast and the uncertainty quantification of the system across different spatial-temporal scales \citep{curry2011climate, delsole2004predictability, edwards1999global, majda2012lessons, majda2018model}.

Nonlinear ordinary or partial differential equations (ODEs or PDEs) were standard techniques to model these complex phenomena. To characterize the multiscale features and the uncertainty in nature, stochastic differential equations (SDEs) have also become popular modeling tools over the past few decades \citep{van1976stochastic, arnold1974stochastic, kloeden1992stochastic, protter2005stochastic}. When applied to DA, these differential equations act as the forecast models. Their short-term statistical prediction, known as the prior distribution, is combined with the available noisy partial observations via the Bayesian inference. The result is called the posterior distribution, which is the solution of DA. It is worth highlighting that approximations are typically incorporated into the development of practical models. These approximations are essential since the perfect knowledge of nature is seldom known. They are also crucial to reducing the computational cost that facilitates the analysis of the model properties and the implementation of DA and ensemble forecasts. Commonly used approaches include the development of physics-based reduced-order models (ROMs) with proper closures \citep{majda2018strategies, carlberg2013gnat, noack2011reduced, taira2020modal, xie2018data,wang2017physics,wu2018physics}, stochastic parameterizations \citep{berner2017stochastic, mana2014toward, dawson2015simulating,schneider2021learning}, and data-driven surrogate models \citep{ahmed2021closures, chekroun2017data, lin2021data, mou2021data, peherstorfer2015dynamic, hijazi2020data, smarra2018data,chen2023operator}.

With the rapid growth of artificial intelligence in recent years, machine learning has become one of the dominant methods for studying complex dynamical systems \citep{cheng2023machine, tang2020introduction, brunton2019data, qian2020lift,duraisamy2019turbulence,brunton2020machine,wu2023learning}, especially for DA and prediction \citep{brajard2020combining, arcucci2021deep, buizza2022data}. On the one hand, machine learning has been widely used as a computationally efficient surrogate of the complicated knowledge-based forecast models \citep{gottwald2021combining, chattopadhyay2023deep, gilbert2010machine, wang2016auto, otto2019linearly, takeishi2017learning} or played the role as a statistical correction to the knowledge-based models that mitigates the model error in the forecast step of DA \citep{gagne2020machine, wikner2021using, rasp2018deep, bonavita2020machine, farchi2021comparison, malartic2022state, farchi2023online, bocquet2020online}. It has also been exploited to optimize the tuning parameters in ensemble DA, such as the inflation rate of the covariance matrix \citep{cheng2022observation, vega2013cello, liu2018deep}. On the other hand, machine learning has been applied more directly by building end-to-end learning schemes for the entire DA system \citep{revach2022kalmannet, boudier2020dan, ouala2018neural, manucharyan2021deep, mou2023combining, boudier2023data}. In terms of learning surrogate models with DA performed, an auto-differentiable DA tool~\citep{chen2022autodifferentiable} has been developed and derivative-free optimization technique~\citep{wu2023learning} has been explored for the ensemble-based DA methods.

In this paper, a new hybrid knowledge-based and machine learning modeling approach, called conditional Gaussian neural stochastic differential equation (CGNSDE), is developed to facilitate modeling complex dynamical systems and implementing the associated DA. The CGNSDE starts by exploiting a systematic causal inference approach to build a simple knowledge-based nonlinear model that nevertheless captures as much explainable physics as possible. Then, neural networks are supplemented to the knowledge-based model, aiming to characterize the remaining features that are challenging to model with simple forms. Notably, the neural networks are combined with the knowledge-based model in a specific way that advances the use of analytic formulae to compute the nonlinear DA solution. In addition to significantly accelerating the computational efficiency when applying the CGNSDE to online DA, these analytic formulae allow us to explicitly augment the standard loss function in training CGNSDE as the predictive model with an additional loss that evaluates the DA skill. The analytic formulae advance a rapid and effective way to incorporate both the path-wise error and the posterior uncertainty into the DA loss.

The CGNSDE has several unique features that distinguish it from many existing machine learning approaches in modeling complex systems and data assimilation. First, the explainable physical components are the critical building blocks of the CGNSDE. To this end, a computationally efficient causal inference via information theory is applied to systematically derive the knowledge-based nonlinear model. The robust model identification process aims to discover the explainable large-scale physics in multiscale complex turbulent systems. In contrast, the machine learning components in the hybrid model can be regarded as the statistical parameterizations of the small-scale or more complicated unresolved features of the underlying system. Second, both the knowledge-based nonlinear model and the entire hybrid model are required to satisfy the so-called conditional Gaussian nonlinear structures. It has been shown that the conditional Gaussian nonlinear structures are ubiquitous in describing or approximating many natural and engineering phenomena \citep{chen2018conditional, chen2022conditional, chen2016filtering}. In addition to being physically consistent with nature, one salient feature of the conditional Gaussian nonlinear structure is that, despite the strong nonlinearity and non-Gaussian statistics, the conditional distribution of the unobserved states given the observations can be written down using closed analytic formulae. Such a conditional distribution is precisely the posterior distribution in DA. As a result, the analytically solvable statistics prevent the use of ensemble methods in DA, enhancing computational efficiency and avoiding empirical tuning to mitigate numerical sampling errors. Third, the efficiency in solving the conditional statistics analytically allows the incorporation of the DA loss into the loss function for training the neural network component, which naturally improves the skill of the CGNSDE in state estimation. Reciprocally, as the DA performance relies on the interdependence between different state variables, such an additional loss will advance the neural network to improve the identification of the causal relationship of the underlying system, further enhancing the modeling skills of the CGNSDE. With the explicit DA loss, the CGNSDE is more capable of estimating extreme events and quantifying the associated uncertainty.

The rest of the paper is organized as follows. The CGNSDE is described in Section \ref{Sec:CGNSDE}. Section \ref{Sec:CGNSDE_Development} presents the procedure of determining the knowledge-based components via causal inference and the training of the CGNSDE by including the DA loss. Section \ref{Sec:Numerics} shows the performance of the CGNSDE via a hierarchy of numerical experiments. The conclusion and discussions are included in Section \ref{Sec:Conclusion}.

\section{Conditional Gaussian Neural Stochastic Differential Equation (CGNSDE)}\label{Sec:CGNSDE}
\subsection{The conditional Gaussian nonlinear system}
The conditional Gaussian nonlinear system (CGNS) is a class of nonlinear and non-Gaussian SDEs, which has wide applications in various disciplines. The general expression of the CGNS is \citep{liptser2013statistics, chen2018conditional}:
\begin{equation}
\begin{aligned}
\label{eq:CGNS}
&\frac{\diff \mathbf{u}_1}{\diff t} = \mathbf{f}_1(\mathbf{u}_1) + \mathbf{g}_1(\mathbf{u}_1)\mathbf{u}_2 + \boldsymbol{\sigma}_1(\mathbf{u}_1)\dot{\mathbf{W}}_1,\\
&\frac{\diff \mathbf{u}_2}{\diff t} = \mathbf{f}_2(\mathbf{u}_1) + \mathbf{g}_2(\mathbf{u}_1)\mathbf{u}_2 + \boldsymbol{\sigma}_2(\mathbf{u}_1)\dot{\mathbf{W}}_2,
\end{aligned}
\end{equation}
where $\mathbf{u}_1\in\mathbb{R}^{N_1}$ and $\mathbf{u}_2\in\mathbb{R}^{N_2}$ are the multi-dimensional state variables considered here. The vectors $\mathbf{f}_1$ and $\mathbf{f}_2$ have the same dimension as $\mathbf{u}_1$ and $\mathbf{u}_2$, respectively, while $\mathbf{g}_1\in\mathbb{R}^{N_1\times N_2}$ and $\mathbf{g}_2\in\mathbb{R}^{N_2\times N_2}$ are two matrices. The two vectors $\dot{\mathbf{W}}_1\in\mathbb{R}^{N'_1}$ and $\dot{\mathbf{W}}_2\in\mathbb{R}^{N'_2}$ are white noises, and the noise coefficients $\boldsymbol{\sigma}_1\in\mathbb{R}^{N_1\times N'_1}$ and $\boldsymbol{\sigma}_2\in\mathbb{R}^{N_2\times N'_2}$ are two matrices. The dimension $N'_1$ of the noise vector $\dot{\mathbf{W}}_1$ does not necessarily equal that of the state variable $\mathbf{u}_1$ (similar for the dimensions of $\mathbf{u}_2$ and $\dot{\mathbf{W}}_2$), though in many applications they are set to be the same for simplicity. All the six vectors and matrices $\mathbf{f}_i$, $\mathbf{g}_i$ and $\boldsymbol{\sigma}_i$, with $i=1,2$, on the right-hand side of \eqref{eq:CGNS} can be any nonlinear functions of $\mathbf{u}_1$ and time $t$, though for notation simplicity the explicit time dependence is omitted. Due to such nonlinear dependence on $\mathbf{u}_1$, the system \eqref{eq:CGNS} is highly nonlinear in terms of the coupled state variables $(\mathbf{u}_1, \mathbf{u}_2)^\mathtt{T}$. As a result, the marginal distributions $p(\mathbf{u}_1)$ and $p(\mathbf{u}_2)$ and joint distribution $p(\mathbf{u}_1,\mathbf{u}_2)$ can all be highly non-Gaussian. Nevertheless, both equations in \eqref{eq:CGNS} depend on $\mathbf{u}_2$ in a conditionally linear way. Therefore, given a trajectory of $\mathbf{u}_1$, the equations in \eqref{eq:CGNS} become a conditional linear system with respect to $\mathbf{u}_2$, and the conditional distribution $p(\mathbf{u}_2(t)|\mathbf{u}_1(s), s\leq t) \sim \mathcal{N}(\boldsymbol{\mu}, \mathbf{R})$ is Gaussian. Such a conditional distribution is the posterior distribution of DA (more precisely the filtering solution), where the trajectory of $\mathbf{u}_1$ up to time $t$, namely $\mathbf{u}_1(s\leq t)$, is the observations while the state of $\mathbf{u}_2$ at $t$ needs to be estimated. Notably, the conditional mean $\boldsymbol{\mu}$ and conditional covariance $\mathbf{R}$ can be solved by the closed analytic formulae:
\begin{equation}
\begin{aligned}
\label{eq:CGNS_Filter}
&\frac{\diff \boldsymbol{\mu}}{\diff t} = (\mathbf{f}_2 + \mathbf{g}_2\boldsymbol{\mu}) + (\mathbf{R}\mathbf{g}_1^\mathtt{T})(\boldsymbol{\sigma}_1 \boldsymbol{\sigma}_1^\mathtt{T})^{-1} \left(\frac{\diff \mathbf{u}_1}{\diff t}-(\mathbf{f}_1 + \mathbf{g}_1\boldsymbol{\mu})\right),\\
&\frac{\diff \mathbf{R}}{\diff t} = \mathbf{g}_2\mathbf{R} + \mathbf{R}\mathbf{g}_2^\mathtt{T} + \boldsymbol{\sigma}_2\boldsymbol{\sigma}_2^\mathtt{T} - \mathbf{R}\mathbf{g}_1^\mathtt{T}(\boldsymbol{\sigma}_1\boldsymbol{\sigma}_1^\mathtt{T})^{-1}(\mathbf{g}_1\mathbf{R}).
\end{aligned}
\end{equation}

Many complex nonlinear dynamical systems fit into the modeling framework \eqref{eq:CGNS}. Some well-known classes of the models are physics-constrained nonlinear stochastic models (for example the noisy versions of Lorenz models, low-order models of Charney-DeVore flows, and a paradigm model for topographic mean flow interaction), stochastically coupled reaction-diffusion models in neuroscience and ecology (for example stochastically coupled FitzHugh-Nagumo models and stochastically coupled SIR epidemic models), and multiscale models for geophysical flows (for example the Boussinesq equations with noise and stochastically forced rotating shallow water equation) \citep{chen2018conditional}. This modeling framework has been exploited to develop realistic systems for the Madden-Julian oscillation and Arctic sea ice \citep{chen2014predicting, chen2022efficient}.

In addition to modeling many natural phenomena, the CGNS framework and its closed analytic DA formulae have been applied to study many theoretical and practical problems. The framework has been utilized to develop a nonlinear Lagrangian DA algorithm, allowing rigorous analysis to study model error and uncertainty \citep{chen2014information, chen2016model, chen2023lagrangian}. The analytically solvable DA scheme has been applied to the state estimation and the prediction of intermittent time series for the monsoon and other climate phenomena \citep{chen2016filtering2, chen2018predicting}. Notably, the efficient DA procedure also helps develop a rapid algorithm to solve high-dimensional Fokker-Planck equation \citep{chen2017beating, chen2018efficient}. The classical Kalman-Bucy filter~\citep{kalman1961new} is the simplest special example of \eqref{eq:CGNS_Filter}.

It is also worth highlighting that the ideas of the CGNS modeling framework and the associated DA procedure have been applied to a much wider range of problems. Examples include developing forecast models in dynamic stochastic superresolution \citep{branicki2013dynamic, keating2012new}, building stochastic superparameterizations for geophysical turbulence \citep{majda2009mathematical, grooms2014stochastic, majda2014new}, and designing efficient multiscale DA schemes \citep{majda2014blended, deng2024lemda}. All these facts indicate that the CGNS provides a valuable building block for many practical methods.

When characterizing multiscale systems, the CGNS is quite effective in modeling large-scale features. Note that the nonlinearity in many applications, especially those in geophysics and fluids, is quadratic, which comes from advection or convection. If $\mathbf{u}_1$ and $\mathbf{u}_2$ are utilized to describe the large- and small-scale state variables, respectively, then the quadratic nonlinearities between $\mathbf{u}_1$ and itself and between $\mathbf{u}_1$ and $\mathbf{u}_2$ can be accurately captured. In some systems, cubic damping appears. CGNS can also involve such a strong damping in the large-scale dynamics. The major nonlinearity that the CGNS cannot fully characterize is the quadratic self-interactions among the small-scale variables, as $\mathbf{u}_2$ is only allowed to appear in a conditional linear way. In such a situation, stochastic parameterizations are often adopted to approximate the self-nonlinear interactions of $\mathbf{u}_2$. A straightforward approach is to replace the quadratic terms of $\mathbf{u}_2$ by stochastic noise and additional damping \citep{majda1999models, majda2001mathematical}, which works well if $\mathbf{u}_2$ lies in the fast time scale. Yet, designing suitable approximate strategies with parsimonious and explicit expressions is generally a nontrivial task. This opens the doors for machine learning to supplement the CGNS.

\subsection{CGNSDE for modeling complex systems and DA}
The CGNSDE is a hybrid knowledge-based and neural network version of the CGNS. It is required to have a similar structure as the CGNS in \eqref{eq:CGNS}, but the functions that depend on $\mathbf{u}_1$ are generalized to involve neural networks. The CGNSDE reads:
\begin{equation}
\begin{aligned}
\label{eq:CGNSDE}
&\frac{\diff \mathbf{u}_1}{\diff t} = \widetilde{\mathbf{f}}_1(\mathbf{u}_1) + \widetilde{\mathbf{g}}_1(\mathbf{u}_1)\mathbf{u}_2 + \boldsymbol{\sigma}_1(\mathbf{u}_1)\dot{\mathbf{W}}_1,\\
&\frac{\diff \mathbf{u}_2}{\diff t} = \widetilde{\mathbf{f}}_2(\mathbf{u}_1) + \widetilde{\mathbf{g}}_2(\mathbf{u}_1)\mathbf{u}_2 + \boldsymbol{\sigma}_2(\mathbf{u}_1)\dot{\mathbf{W}}_2,
\end{aligned}
\end{equation}
where the functions with the terms with tildes represent the combinations of knowledge-based terms and neural networks that only depend on $\mathbf{u}_1$. More specifically, the CGNSDE in \eqref{eq:CGNSDE} can be rewritten into the following more detailed form:
\begin{equation}
\begin{aligned}
\label{eq:CGNSDE_Details}
&\frac{\diff \mathbf{u}_1}{\diff t} = \mathbf{f}_1(\mathbf{u}_1) + \mathbf{g}_1(\mathbf{u}_1)\mathbf{u}_2 + \mathbf{NN}_{1,1}(\mathbf{u}_1)+ \mathbf{NN}_{1,2}(\mathbf{u}_1)\mathbf{u}_2 + \boldsymbol{\sigma}_1(\mathbf{u}_1)\dot{\mathbf{W}}_1,\\
&\frac{\diff \mathbf{u}_2}{\diff t} = \mathbf{f}_2(\mathbf{u}_1) + \mathbf{g}_2(\mathbf{u}_1)\mathbf{u}_2 + \mathbf{NN}_{2,1}(\mathbf{u}_1)+ \mathbf{NN}_{2,2}(\mathbf{u}_1)\mathbf{u}_2 + \boldsymbol{\sigma}_2(\mathbf{u}_1)\dot{\mathbf{W}}_2,
\end{aligned}
\end{equation}
where $\mathbf{f}_1, \mathbf{g}_1, \mathbf{f}_2$ and $\mathbf{g}_2$ are nonlinear functions of $\mathbf{u}_1$ with explainable forms (e.g., a nonlinear combination of some candidate functions), while $\mathbf{NN}_{i,j}$ with $i,j=1,2$ are neural network functions to further enhance the ability of the CGNSDE in capturing the variability of nature. The neural network components $\mathbf{NN}_{i,j}$, can be formalized using four distinct parts derived either from the output of a single neural network or from several neural networks. Note that the diffusion coefficients $\boldsymbol{\sigma}_1$ and $\boldsymbol{\sigma}_2$ in the CGNSDE can, in principle, be generalized as a combination of explainable functions and neural networks as well. Nevertheless, they are modeled by only the parametric forms in this work for simplicity. One key feature of \eqref{eq:CGNSDE_Details} is that those neural network components only allow $\mathbf{u}_1$ as the input, which guarantees the conditional Gaussianity of the entire system. It is worth noting that the input of the neural network part does not necessarily have to be the value of $\mathbf{u}_1$ at the current time $t$. It can consist of a segment of the trajectory of $\{\mathbf{u}_1(s)|s\leq t\}$. Input with such non-Markovian terms nevertheless preserves the conditional Gaussianity of the system. This feature allows a significant degree of freedom to the design of neural networks, allowing both the feed-forward and recurrent neural networks to apply to the CGNSDE. With the dependence of $\mathbf{f}_1, \mathbf{g}_1, \mathbf{f}_2$ and $\mathbf{g}_2$ only on $\mathbf{u}_1$ in the CGNSDE, the conditional distribution $p(\mathbf{u}_2(t)|\mathbf{u}_1(s), s\leq t) \sim \mathcal{N}(\boldsymbol{\mu}, \mathbf{R})$, which is the posterior distribution of DA, is Gaussian and its time evolution can be written down using analytic formulae that are analogs to \eqref{eq:CGNS_Filter}.

The CGNSDE has several unique advantages. First, the explainable physical components are the critical building blocks of the CGNSDE. This distinguishes the CGNSDE from the non-parametric models that replace the entire right-hand side of the original system with neural networks. These are helpful predictive models. However, these models are hard to be used to reveal the underlying physics. Such pure neural-network-based models are also challenging to effectively use for solving inverse problems, such as DA. With the large-scale features captured by explainable physical components, the neural networks play a more critical role in characterizing the residual. Therefore, they share many essential features as the residual networks that advance the performance of the neural networks in more efficiently capturing the features of the underlying dynamics \citep{he2016deep, he2015delving}. Second, the analytically solvable posterior distribution prevents the use of ensemble methods in DA. These analytic formulae not only enhance computational efficiency but also avoid empirical tunings in the standard ensemble DA that is essential to mitigate numerical sampling errors. Third, the efficient DA solver with these analytic formulae allows the incorporation of the DA loss into the target loss function to train the CGNSDE, which naturally improves the skill for state estimation. Reciprocally, as the performance of DA depends on the accuracy in modeling the interdependence between different state variables, namely their causal relationship, incorporating the DA loss will further enhance the performance of the CGNSDE in modeling the underlying dynamical features. With the explicit DA loss, the CGNSDE is also more capable of estimating extreme events and quantifying the associated uncertainty.

\section{Procedure of Developing a CGNSDE}
\label{Sec:CGNSDE_Development}
Assume that a sufficiently long time series of both $\mathbf{u}_1$ and $\mathbf{u}_2$ are available in the model development stage to determine the CGNSDE.
In the testing stage for the online DA and forecast, only the observations of $\mathbf{u}_1$ are needed, which is consistent with the realistic situations with partial observations.

The development of the CGNSDE involves two steps. First, the explicit expressions of the terms $\mathbf{f}_1(\mathbf{u}_1) + \mathbf{g}_1(\mathbf{u}_1)\mathbf{u}_2$ and $\mathbf{f}_2(\mathbf{u}_1) +\mathbf{g}_2(\mathbf{u}_1)\mathbf{u}_2$ need to be determined to include as much physically explainable information as possible. Second, suitable architectures and loss functions are designed for the neural networks $\mathbf{NN}_{i,j}$ with $i,j=1,2$. Notably, the order of the two steps is essential to prevent the neural networks from taking away the information described by the physically explainable components.

\subsection{Determining the knowledge-based components by causal inference}
\label{ssec:CGNSDE_Step1}

The knowledge-based components in the CGNSDE contain the terms $\mathbf{f}_1(\mathbf{u}_1) + \mathbf{g}_1(\mathbf{u}_1)\mathbf{u}_2$ and $\mathbf{f}_2(\mathbf{u}_1) +\mathbf{g}_2(\mathbf{u}_1)\mathbf{u}_2$ in \eqref{eq:CGNSDE_Details}. These functions are often determined by human knowledge case by case. Yet, in a more general situation, an automatic learning algorithm is preferred to determine these terms systematically. To this end, a causal inference method based on the so-called causation entropy is adopted to achieve such a goal \citep{almomani2020entropic, almomani2020erfit, elinger2021causation,chen2023causality,chen2023ceboosting}. Causation entropy is an efficient machine-learning method for the sparse identification of dynamical systems. The causal relationship facilitates the discovery of the underlying explainable knowledge-based components of the system. Notably, the model identification results using the causation entropy method are more robust to noise or chaotic features than the standard least absolute shrinkage and selection operator (LASSO) regressions. It has been shown that in the presence of even slight random noise, both the covariate selection accuracy and the fraction of zero entries may decrease significantly \citep{elinger2020information} when applying the standard LASSO regression.

The procedure of applying the causation entropy in determining the physically explainable components of the system is as follows.\medskip

\noindent\textbf{Step 1. Determining the state variables.}
The state variables of the CGNSDE are pre-determined. To facilitate the presentation, these variables are included into an $N$-dimensional column vector:
\begin{equation}
\mathbf{U}=(\mathbf{u}_1, \mathbf{u}_2)^\mathtt{T}=(u_1,\ldots, u_{N_1}, u_{N_1+1},\ldots, u_{N_1+N_2})^{\mathtt{T}},
\end{equation}
where $N=N_1+N_2$.\medskip

\noindent\textbf{Step 2. Developing a function library.}
After determining the state variables, a library $\mathbf{h}$ consisting of in total $M$ possible candidate functions to describe the right-hand side of the model is developed,
\begin{equation}\label{eq:Library}
  \mathbf{h} = \{h_1,\ldots, h_{m-1}, h_m, h_{m+1}, \ldots, h_M\}.
\end{equation}
Typically, a large number of candidate functions is included in the library to allow for coverage over different possible dynamical features of the underlying true dynamics. Each $h_m$ is given by a linear or nonlinear function containing a few components of $\mathbf{U}$. To follow the structure of CGNS in \eqref{eq:CGNS}, the components within $\mathbf{u}_2$ are required to be linear in the candidate function $h_m$. Prior knowledge of the dynamics can help determine the function library. The library can also include as many potential candidate functions as possible provided that the elements within $\mathbf{u}_2$ are incorporated linearly, allowing the causal inference to determine the useful ones automatically.\medskip

\noindent\textbf{Step 3. Computing the causation entropy.}
Next, causal inference is utilized to determine the model structure. To this end, a causation entropy $C_{h_{m} \rightarrow \dot{u}_i \mid\left[\mathbf{h} \backslash {h}_{m}\right]}$ is computed to detect if the candidate function $h_m$ contributes to the right-hand side of the equation for $u_i$, namely $\d u_i/\d t :=\dot{u}_i$. The causation entropy is given by \citep{almomani2020entropic, almomani2020erfit, elinger2021causation}:
\begin{equation}\label{Causation_Entropy}
  C_{h_{m} \rightarrow \dot{u}_i \mid\left[\mathbf{h} \backslash {h}_{m}\right]}=H(\dot{u}_i|\left[\mathbf{h} \backslash {h}_{m}\right]) - H(\dot{u}_i|\mathbf{h}),
\end{equation}
where $\mathbf{h} \backslash {h}_{m}$ represent the set that contains all functions in $\mathbf{h}$ except $h_m$. In other words, $\mathbf{h} \backslash {h}_{m}$ contains $M-1$ candidate functions and is defined as
\begin{equation}\label{Library2}
  \mathbf{h}\backslash {h}_{m} = \{h_1,\ldots, h_{m-1},h_{m+1},\ldots, h_M\}.
\end{equation}
The term $H(\cdot|\cdot)$ is the conditional entropy, which is related to Shannon's entropy $H(\cdot)$ and the joint entropy $H(\cdot,
\cdot)$. For two multi-dimensional random variables $\mathbf{X}$ and $\mathbf{Y}$ (with the corresponding states being $\mathbf{x}$ and $\mathbf{y}$), they are defined as \citep{cover1999elements}:
\begin{equation}\label{Entropies}
\begin{split}
  H(\mathbf{X}) &= -\int_x p(\mathbf{x})\log(p(\mathbf{x}))\d \mathbf{x},\\
  H(\mathbf{Y}| \mathbf{X}) &= -\int_\mathbf{x}\int_\mathbf{y} p(\mathbf{x},\mathbf{y})\log(p(\mathbf{y}|\mathbf{x}))\d \mathbf{y}\d \mathbf{x},\\
  H(\mathbf{X},\mathbf{Y}) &= -\int_\mathbf{x}\int_\mathbf{y} p(\mathbf{x},\mathbf{y})\log(p(\mathbf{x},\mathbf{y}))\d \mathbf{y}\d \mathbf{x},
\end{split}
\end{equation}
where $p(\mathbf{x})$ is the PDF of $\mathbf{x}$ and $p(\mathbf{y}|\mathbf{x})$ is the conditional PDF of $\mathbf{y}$ on $\mathbf{x}$. On the right-hand side of \eqref{Causation_Entropy}, the difference between the two conditional entropies indicates the information in $\dot{u}_i$ contributed by the specific function $h_m$ given the contributions from all the other functions. Thus, it tells if $h_m$ provides additional information to $\dot{u}_i$ conditioned on the other potential terms in the dynamics. It is worthwhile to highlight that the causation entropy in \eqref{Causation_Entropy} is fundamentally different from directly computing the correlation between $\dot{u}_i$ and $h_m$, as the causation entropy also considers the influence of the other library functions. If both $\dot{u}_i$ and $h_m$ are caused by a common factor $h_{m^\prime}$, then $\dot{u}_i$ and $h_m$  can be highly correlated. Yet, in such a case, the causation entropy $C_{h_{m} \rightarrow \dot{u}_i \mid\left[\mathbf{h} \backslash {h}_{m}\right]}$ will be zero as $h_m$ is not the causation of $\dot{u}_i$.

The causation entropy is computed from each of the candidate functions in $\mathbf{h}$ to each $\dot{{u}}_i$. Thus, there are in total $N\times M$ causation entropies, which can be written as a $N\times M$ matrix, called the causation entropy matrix. Note that the dimension $\mathbf{X}$ in \eqref{Entropies} is $M$ when it is applied to compute the second term on the right-hand side of the causation entropy in \eqref{Causation_Entropy}. This implies that the direct calculation of the entropies in \eqref{Entropies} involves a high-dimensional numerical integration, which is a well-known computationally challenging issue \citep{bellman1961dynamic}. To circumvent the direct numerical integration, the entropy calculation approximates all the joint and marginal distributions as Gaussian. In such a way, the causation entropy can be computed by
\begin{equation}
\label{Entropy_Gaussians}
\begin{split}
C_{\mathbf{Z} \rightarrow \mathbf{X} | \mathbf{Y}} &=H(\mathbf{X} | \mathbf{Y})-H(\mathbf{X} | \mathbf{Y}, \mathbf{Z}) \\
& = H(\mathbf{X},\mathbf{Y}) - H(\mathbf{Y}) - H(\mathbf{X},\mathbf{Y},\mathbf{Z}) + H(\mathbf{Y},\mathbf{Z})\\
& \approx \frac{1}{2} \ln(\operatorname{det}(\mathbf{R}_{\mathbf{X}\mathbf{Y}}))-\frac{1}{2} \ln(\operatorname{det}(\mathbf{R}_{\mathbf{Y}})) - \frac{1}{2} \ln(\operatorname{det}(\mathbf{R}_{\mathbf{X}\mathbf{Y}\mathbf{Z}}))
 +\frac{1}{2} \ln(\operatorname{det}(\mathbf{R}_{\mathbf{Y}\mathbf{Z}})),
\end{split}
\end{equation}
where $\mathbf{R}_{\mathbf{XYZ}}$ denotes the covariance matrix of the state variables $(\mathbf{X},\mathbf{Y},\mathbf{Z})$ and similar for other covariance. The notations $\ln(\cdot)$ and $\operatorname{det}(\cdot)$ are the logarithm of a number and the determinant of a matrix, respectively.

The simple and explicit expression in \eqref{Entropy_Gaussians} based on the Gaussian approximation can efficiently compute the causation entropy. It allows the computation of the causation entropy with a moderately large dimension, sufficient for many applications. It is worth noting that the Gaussian approximation may lead to certain errors in computing the causation entropy if the true distribution is highly non-Gaussian. Nevertheless, the primary goal is not to obtain the exact value of the causation entropy. Instead, it suffices to detect if the causation entropy $C_{h_{m} \rightarrow \dot{u}_i \mid\left[\mathbf{h} \backslash {h}_{m}\right]}$ is nonzero (or practically above a small threshold value). In most applications, if a significant causal relationship is detected in the higher-order moments, it is very likely in the Gaussian approximation. This allows us to efficiently determine the sparse model structure, where the exact values of the nonzero coefficients on the right-hand side of the model will be calculated via a simple maximum likelihood estimation to be discussed in the following. Note that the Gaussian approximation has been widely applied to compute various information measurements and leads to reasonably accurate results \citep{majda2018model, tippett2004measuring, kleeman2011information, branicki2012quantifying, branicki2013non}.

With the $N\times M$ causation entropy matrix in hand, the next step is determining the model structure. This can be done by setting up a threshold value of the causation entropy and retaining only those candidate functions with the causation entropies exceeding the threshold. This will exclude those terms with small values of causation entropy, which is usually due to numerical error using a finite time series. The resulting model will contain only functions that significantly contribute to the dynamics and facilitate a sparse model structure. Sparsity is crucial to discovering the correct underlying physics and prevents overfitting \citep{ying2019overview, brunton2016discovering}. It will also guarantee the robustness of the model in response to perturbations and allow the model to apply to certain extrapolation tests. \medskip

\noindent\textbf{Step 4. Parameter estimation.}
The final step is to estimate the parameters in the resulting model. Despite the nonlinearity in the underlying dynamics, the parameters usually appear in a linear way. In such a case, parameter estimation can be easily handled using a simple regression method or a maximum likelihood estimator. See \citep{chen2020learning} for the technical details. Notably, closed analytic formulae are available, making the procedure efficient and accurate. It is worth highlighting that constraints to the parameter values are often included in the parameter estimation procedure. In many turbulent systems, the quadratic nonlinear terms are assumed to be energy conserved. This is motivated by most geophysical systems where the quadratic nonlinearity is the advection and is a natural conservation quantity. The energy-conserving quadratic nonlinearity prevents the finite time blowup of the solution in the derived model and is physically consistent \citep{majda2012physics, harlim2014ensemble}. Remarkably, closed analytic formulae are still available for parameter estimation in the presence of such constraints.

\subsection{Determining the neural network components}
\label{ssec:CGNSDE_Step2}

Although Step 4 in Section \ref{ssec:CGNSDE_Step1} provides a direct way to estimate the parameters in the knowledge-based components $\mathbf{f}_1(\mathbf{u}_1), \mathbf{g}_1(\mathbf{u}_1)\mathbf{u}_2, \mathbf{f}_2(\mathbf{u}_1), \mathbf{g}_2(\mathbf{u}_1)\mathbf{u}_2$, the parameters of these physically explainable terms in the CGNSDE are estimated in a slightly different way. Specifically, these parameters are not immediately estimated after the model format is determined by the causal inference. Once the model structure of the physically explainable components is determined (Steps 1-3 in Section \ref{ssec:CGNSDE_Step1}), the neural network and the parameters of the physically explainable components are determined by a joint learning algorithm. This allows a simultaneous optimization of the two components in the CGNSDE.

It is worth noting that the CGNSDE in \eqref{eq:CGNSDE_Details} is implemented to support automatic differentiation, which facilitates the training process using gradient descent methods. Training the neural network components in the CGNSDE involves using a standard forecast loss and a DA loss. The latter plays a vital role in improving the skillful DA using the CGNSDE. Notably, the closed analytic formulae of the DA solution facilitate the incorporation of such an additional crucial loss function.

\subsubsection{The short-term forecast loss}
\label{ssec:CGNSDE_forecast_loss}
Denote by $\Tilde{\mathbf{U}}(t_n)=(\Tilde{\mathbf{u}}_1(t_n), \Tilde{\mathbf{u}}_2(t_n))^\mathtt{T}$, for $n=1,\ldots,N_s$, which is a multi-dimensional series predicted by the CGNSDE \eqref{eq:CGNSDE_Details} starting from $\mathbf{U}(t_0)$. The hyper-parameter $N_s$, which determines the forecast horizon, is usually a relatively small integer since the path-wise forecast diverges quickly for chaotic systems.

The forecast loss $L_{\textrm{forecast}}$ is defined as the averaged error between the predicted states $\{ \Tilde{\mathbf{U}}(t_n) \}_{n=1}^{N_s}$ and the truth $\{ \mathbf{U}(t_n) \}_{n=1}^{N_s}$:
\begin{equation}
\label{eq:forecast_mse}
    L_{\textrm{forecast}}(\mathbf{U}, \Tilde{\mathbf{U}}):= \frac{1}{N_s}\sum_{n=1}^{N_s}  \| \mathbf{U}(t_n) - \Tilde{\mathbf{U}}(t_n) \|^2,
\end{equation}
where $\|\cdot\|$ is the standard vector $\ell^2$-norm.

In each training epoch, a time series of $\mathbf{U}$ with $N_s$ time steps is randomly sampled from the training data as the truth. Starting from the same initial value as the truth, the CGNSDE will be integrated for $N_s$ steps to obtain the predicted states. Utilizing an auto-differentiation engine, the neural networks can be trained using the stochastic gradient descent (SGD) algorithm. In this work, an epoch is defined as a single iteration in which SGD is executed using a selected batch.

\subsubsection{The DA loss}
\label{ssec:CGNSDE_DA_loss}
The DA loss assesses the error in the DA solution computed from the CGNSDE. Notably, the explicit formulae in \eqref{eq:CGNS_Filter} are used to facilitate the calculation of the DA loss.

One pre-requite of carrying out the DA is to determine the noise coefficients $\boldsymbol{\sigma}_1$ and $\boldsymbol{\sigma}_2$ in \eqref{eq:CGNSDE_Details}.
For the simplicity of presenting the idea, hereafter the dimensions of the white noise $\dot{\bf W}_1$ and $\dot{\bf W}_2$ in \eqref{eq:CGNSDE_Details} are assumed to be the same as the state $\mathbf{u}_1$ and $\mathbf{u}_2$, respectively, e.g., $N'_1=N_1$ and $N'_2=N_2$. Further assume $\boldsymbol{\sigma}_1$ and $\boldsymbol{\sigma}_2$ are diagonal positive definite matrix. With a pre-trained CGNSDE by only using the forecast loss, the diagonal elements in $\boldsymbol{\sigma}_1$ and $\boldsymbol{\sigma}_2$ can be estimated by the quadratic variation:
\begin{equation}
    \begin{aligned}
        \mathrm{diag}(\boldsymbol{\sigma}_i) = \sqrt{ \frac{\Delta t}{N_t}\sum_{n=1}^{N_t} \bigl(\dot{\mathbf{u}}_i(t_n) -  \Tilde{\dot{\mathbf{u}}}_i(t_n)\bigr) \odot \bigl(\dot{\mathbf{u}}_i(t_n) -  \Tilde{\dot{\mathbf{u}}}_i(t_n)\bigr)}, \qquad \mbox{for~} i = 1, 2,
    \end{aligned}
    \label{eq:sigma_estimation}
\end{equation}
where $\Delta t$ is the numerical integration time step, $N_t$ is the total number of the time steps in the training data, and the notation $\odot$ denotes the element-wise product. In \eqref{eq:sigma_estimation}, the time derivative of the true signal, namely $\dot{\mathbf{u}}_i(t_n)$, can often be obtained by calculating the numerical derivatives of the true state $\mathbf{u}_i$ at time $t_n$, while $\Tilde{\dot{\mathbf{u}}}_i(t_n)$ is the time derivative of the corresponding prediction from the pre-trained CGNSDE.
With the estimated $\boldsymbol{\sigma}_i$, the posterior mean ${\boldsymbol{\mu}}$ and covariance $\mathbf{R}$ of the unobserved variables $\mathbf{u}_2$ in CGNSDE can be calculated from the closed analytic formulae in \eqref{eq:CGNS_Filter}, given the time series of the observed variables $\mathbf{u}_1$.

One natural way of constructing a DA loss $L_{\textrm{DA}}$ is to compute the difference between the posterior mean estimate $\boldsymbol{\mu}$ and the true signal $\mathbf{u}_2$:
\begin{equation}
    \label{eq:DA_MSE}
    \begin{aligned}
        L_{\textrm{DA}}(\mathbf{u}_2, \boldsymbol{\mu}) := \frac{1}{N_l-N_b}\sum_{n=N_b+1}^{N_l} \|\mathbf{u}_2(t_n) - \boldsymbol{\mu}(t_n)\|^2,
    \end{aligned}
\end{equation}
where $N_l$ denotes the number of steps to compute the DA solution during the training period. Note that $N_l$ is usually much larger than $N_s$ in \eqref{eq:forecast_mse}. This is because the DA solution will not have a quick diverge as the forecast one. Note that although it is natural to set the forecast initial value to be the same as the truth to eliminate the initial inconsistency in the forecast loss, the exact initial distribution of the DA solution (especially the initial uncertainty) is usually unknown. Therefore, it takes some time for the DA solution to be adjusted to eliminate the inconsistency from the initialization. To this end, the few steps of DA, which correspond to the burn-in period, are omitted, which leads to the DA loss being computed from the $(N_b+1)$-th step.

It is worth highlighting that the DA loss in \eqref{eq:DA_MSE} has a straightforward form and can be easily applied in practice. Although only the posterior mean is explicitly adopted in the loss function, the entire posterior information from DA is exploited since the posterior mean depends on the posterior covariance, as can be seen in \eqref{eq:CGNS_Filter}. The DA loss in \eqref{eq:DA_MSE} will be adopted in all the numerical experiments in Section \ref{Sec:Numerics}. The results there will demonstrate that the advantage of including this DA loss in training the CGNSDE to improve the modeling and DA skills, including performing more stable long-term simulations and better capturing the true system behaviors (e.g., the critical statistical properties, the chaotic patterns, and the extreme events). \medskip

\noindent\textbf{An alternative DA loss}.\\
One potential shortcoming of the path-wise DA loss in \eqref{eq:DA_MSE} is that it does not fully use the probabilistic features in the state estimation via DA. Therefore, an alternative DA loss $L_{\textrm{DA}}$ can be defined from a more probabilistic perspective based on the log-likelihood:
\begin{equation}
\begin{aligned}
    \label{eq:DA_NLL}
    L_{\textrm{DA}}(\mathbf{u}_2, \boldsymbol{\mu}, \mathbf{R}) &:= -\ln\Bigl(\prod_{n=N_b+1}^{N_l} p\bigl(\mathbf{u}_2(t_n)|\mathbf{u}_1(s), s \leq t_n\bigr)\Bigr) \\
    &= \frac{1}{2} \sum_{n=N_b+1}^{N_l} \Bigl(N_2\log(2\pi) + \ln(\operatorname{det}(\mathbf{R}(t_n))) + \|\mathbf{u}_2(t_n) - \boldsymbol{\mu}(t_n)\|^2_{\mathbf{R}(t_n)}\Bigr),
\end{aligned}
\end{equation}
where $\|\cdot\|^2_{\mathbf{R}}$ denotes a weighted $\ell^2$-norm, i.e., $\|\boldsymbol{\mu}\|^2_{\mathbf{R}}=\boldsymbol{\mu}^\mathtt{T}\mathbf{R}^{-1}\boldsymbol{\mu}$. The second line in \eqref{eq:DA_NLL} utilizes the fact that $p(\mathbf{u}_2(t_n)|\mathbf{u}_1(s), s \leq t_n)$ follows a Gaussian distribution $\mathcal{N}(\boldsymbol{\mu}(t_n),\mathbf{R}(t_n))$, whose mean $\boldsymbol{\mu}(t_n)$ and covariance matrix $\mathbf{R}(t_n)$ can be obtained by simulating \eqref{eq:CGNS_Filter}. Considering that $N_2\log(2\pi)$ is just a constant factor for any given dynamical system, the main difference between \eqref{eq:DA_NLL} and \eqref{eq:DA_MSE} is that the DA loss based on log-likelihood leads to a weighted $\ell^2$-norm, which allows using the covariance matrix $\mathbf{R}$ to normalizes the mismatch between $\mathbf{u}_2$ and $\boldsymbol{\mu}$. Building a loss function in such a way potentially enhances the importance of the posterior uncertainty in training the CGNSDE. Consequently, the DA skill in recovering intermittency and extreme events is expected to be improved with such a loss function.


\subsubsection{Training the CGNSDE with both the forecast and the DA losses}
\label{ssec:CGNSDE_total_loss}
Given the expression of the short-term forecast loss \eqref{eq:forecast_mse} and the DA loss \eqref{eq:DA_MSE}, the overall target loss function can be defined as a weighted sum of these two functions:
\begin{equation}
\label{eq:total_MSE}
    L_{\textrm{total}}:=\lambda_1 \underbrace{\frac{1}{N_s}\sum_{n=1}^{N_s} \|\mathbf{U}(t_n) -  \Tilde{\mathbf{U}}(t_n)\|^2}_{L_\textrm{forecast}} + \lambda_2  \underbrace{\frac{1}{N_l-N_b}\sum_{j=N_b+1}^{N_l} \|\mathbf{u}_2(t_j) - \boldsymbol{\mu}(t_j)\|^2}_{L_\textrm{DA}},
\end{equation}
where the two constants $\lambda_1$ and $\lambda_2$ are the weights of the forecast loss and DA loss, respectively. In the following numerical experiments, the two weights are assigned as $\lambda_1 = 1/N$, which is the dimension of all the state variables $\mathbf{U}$, and $\lambda_2 = 1/N_2$, which is the dimension of unobserved state variables $\mathbf{u}_2$. With such a choice, $L_{\textrm{forecast}}$ and $L_{\textrm{DA}}$ become the average of the mean squared errors between the truth and the predictions and the state estimation.

In each training epoch, a time series of $\mathbf{U}$ with $N_s$ steps and another time series of $\mathbf{u}_2$ with $N_l$ steps are sampled from the training data. The numerical simulation of $\tilde{\mathbf{U}}$ and $\boldsymbol{\mu}$ are performed based on the CGNSDE solvers that support auto-differentiation. These time series are used to evaluate the total loss defined in \eqref{eq:total_MSE}, with which the auto-differentiation provides the gradient information that can be employed to optimize the unknown parameters in the CGNSDE via gradient descent methods. Same as in Section \ref{ssec:CGNSDE_forecast_loss}, an epoch refers to a single iteration where SGD is executed with a selected batch.

\section{Numerical Results}
\label{Sec:Numerics}
\subsection{Overview}
In all the numerical experiments, the results from the following three models are compared to demonstrate the skillful performance of the CGNSDE. Without loss of generality, the feed-forward network is utilized in all the experiments. These simulations aim to show the necessity of supplementing the neural network components into the knowledge-based models and the crucial role of incorporating the DA loss into the overall loss function.
The three models used below are:
\begin{enumerate}
    \item The knowledge-based regression model. It corresponds to CGNSDE in \eqref{eq:CGNSDE_Details} without the neural network parts. The unknown functions in this model are constructed based on a library of candidate functions and calibrated via a causation-entropy-based system identification method. The detailed procedures are described in Section \ref{ssec:CGNSDE_Step1} from Step 1 to Step 4. Alternatively, the prior knowledge from the users can also be exploited to build such a model, in which case the prior knowledge replaces the causal inference for the model development.
    \item The CGNSDE without DA loss. It corresponds to the CGNSDE model in the form of \eqref{eq:CGNSDE_Details} combining knowledge-based and neural network components. However, only the forecast loss in \eqref{eq:forecast_mse} is utilized to train the neural networks. The detailed procedures to develop this model are Step 1 to Step 3 in Section \ref{ssec:CGNSDE_Step1} and the forecast loss in Section \ref{ssec:CGNSDE_forecast_loss}.
    \item The CGNSDE with the DA loss. It corresponds to the CGNSDE model in the form of \eqref{eq:CGNSDE_Details} and is trained with the total loss function \eqref{eq:total_MSE} that combines both forecast loss and DA loss. The detailed procedures to develop this model are Step 1 to Step 3 in Section \ref{ssec:CGNSDE_Step1} and the total loss in Section \ref{ssec:CGNSDE_total_loss}.
\end{enumerate}

The performance of all three models are examined utilizing the following validation metrics:
\begin{enumerate}
  \item [(a).] Forecast MSE. It is the MSE  \eqref{eq:forecast_mse} between the predicted state $\Tilde{\mathbf{U}}$ and the truth $\mathbf{U}$.
  \item [(b).] DA MSE. It is the DA MSE in \eqref{eq:DA_MSE} between the posterior mean estimate $\boldsymbol{\mu}$ and the unobserved true state $\mathbf{u}_2$.
  \item [(c).] DA negative log-likelihood (Neg-Log-Likelihood). It is the DA negative log-likelihood in \eqref{eq:DA_NLL} calculated based on the true value of the unobserved state $\mathbf{u}_2$ related to the Gaussian distribution built upon the posterior mean $\boldsymbol{\mu}$ and the posterior covariance $\mathbf{R}$. Note that the DA loss in training the neural networks is still based on the error in the posterior mean related to the truth using \eqref{eq:DA_MSE}. The negative log-likelihood \eqref{eq:DA_NLL} based on the entire posterior distribution is only used here as a validation criterion.
\end{enumerate}
It is worth highlighting that the negative log-likelihood function utilizes the direction information from both the posterior mean and the posterior covariance (i.e., the uncertainty). Although the negative log-likelihood function is not directly used as the training loss, it is exploited here as one additional validation metric to assess the DA skill of these models beyond using traditional path-wise measurements such as the MSE.

To demonstrate the performance of CGNSDE on various types of complex dynamical systems, three dynamical systems are utilized to generate the true signal for building and training the CGNSDE. Depending on the choice of the observed and unobserved state variables, the true system can be a CGNS or a non-CGNS. Here non-CGNS refers to any dynamical system that can not be written in the general form of \eqref{eq:CGNS}. On the one hand, when the underlying system is a CGNS, then closed analytic formulae \eqref{eq:CGNS_Filter} can be exploited to compute the exact DA solution. Such a solution, as the optimal estimate for the state, will be compared with the ones from the CGNSDE models and the knowledge-based regression model. On the other hand, if the true system is not a CGNS, then the ensemble Kalman-Bucy filter (EnKBF) \citep{bergemann2012ensemble} will be utilized to provide an approximate optimal reference solution in assessing the DA performance of all the models.

A summary of the true underlying system and the conclusion of the associated numerical experiments is as follows.
\begin{enumerate}
\item The noisy Lorenz 84 system. It is a low-dimensional chaotic system that can be formalized as a CGNS. The study in Section \ref{ssec:results_L84} is utilized as a proof-of-concept of the skillful performance of the CGNSDE. By specifically prescribing imperfect knowledge-based components, the result shows that the additional neural network components significantly improve the model performance, and the DA loss is crucial in enhancing the DA skill.
\item The projected stochastic Burgers-Sivashinsky equation. It is a low-dimensional chaotic system that violates the assumption of CGNS and is featured by extreme events. The results in Section \ref{ssec:results_BSE} show that the CGNSDE with the DA loss can accurately describe the model features and find the DA solution for such a non-CGNS. Notably, the CGNSDE with the DA loss reaches a posterior mean estimation comparable to the one by applying the EnKBF to the true system. The posterior covariance from the DA associated with the CGNSDE accurately characterizes the uncertainty. In addition, the extreme events are well captured by the CGNSDE in both the forecast and DA solutions.
\item The Lorenz 96 system. It is a moderate-dimensional chaotic system with local nonlinear interactions. An effective way of designing and training the low-dimensional components of the neural network.
    \begin{itemize}
    \item Case 1: Homogeneous solution and CGNS. This case is designed to have a statistically homogeneous solution. By specifying the observed and unobserved variables, the system is formalized as a CGNS. The results in Section \ref{sssec:results_L96_1} show that the proposed CGNSDE framework has a good performance and scales well with the dimensionality of the system. In addition, the CGNSDE model trained with the DA loss can provide stable long-term simulations that have comparable statistics with the true system.
    \item Case 2: Homogeneous solution and non-CGNS: By choosing a different set of observed variables, the true system becomes a non-CGNS. The results in Section \ref{sssec:results_L96_2} confirm that the CGNSDE model trained with the DA loss can still handle the modeling and DA for a non-CGNS. The DA results are also comparable to the ones from true system with EnKBF.
    \item Case 3: Inhomogeneous solution and non-CGNS: The parameters are chosen to be spatial-dependent and therefore the solution demonstrates statistically inhomogeneous features. The results in Section \ref{sssec:results_L96_3} show that the CGNSDE model trained with the DA loss can still capture the true hidden states well. More specifically, the mean estimation from CGNSDE with DA has a good agreement with most system states, and the uncertainties are noticeably larger for those states with less satisfactory estimation of their means.
\end{itemize}
\end{enumerate}

\subsection{The Lorenz 84 model: a low-order chaotic system}
\label{ssec:results_L84}
The noisy Lorenz 84 model is a simple analogue of the global atmospheric circulation \citep{vallis2017atmospheric, salmon1998lectures}, which has the following form \citep{lorenz1984formulation, lorenz1984irregularity}:
\begin{equation}
\begin{aligned}
    \label{eq:L84}
    &\frac{\diff x}{\diff t} = -(y^2 + z^2) - a(x-f) + \sigma_x \dot{W}_x,\\
    &\frac{\diff y}{\diff t} = -bxz + xy - y + g + \sigma_y\dot{W}_y,\\
    &\frac{\diff z}{\diff t} = bxy + xz - z + \sigma_z\dot{W}_z.
\end{aligned}
\end{equation}
In \eqref{eq:L84}, the zonal flow $x$ represents the intensity of the mid-latitude westerly wind current. There is a wave component in the system, with $y$ and $z$ denoting the cosine and sine phases, respectively, of a sequence of vortices superimposed on the zonal flow. The wave variables are scaled in relation to zonal flow such that $x^2 + y^2 + z^2$ is the total scaled energy encompassing kinetic, potential, and internal energy components. Note that these equations can be derived as a Galerkin truncation of the two-layer quasigeostrophic potential vorticity equations in a channel.

Viscous and thermal processes will linearly damp the vortices in this system. The parameter $a<1$ is a Prandtl number and the time unit of the system is defined by the damping time. The term $af$, proportional to the contrast between solar heating at low and high latitudes, is the external force that drives the zonal flow. The terms $xy$ and $xz$ illustrate how the wave is amplified through its interaction with the zonal flow. With the wave transporting heat poleward, the temperature gradient will be reduced at a rate proportional to the square of the amplitudes indicated by the term $-(y^2 + z^2)$. If $x>0$, the terms $-bxz$ and $bxy$ represent the wave westward displacement by the zonal current. With $b>1$, the displacement is allowed to overcome the amplification. A secondary forcing $g$ mimics the contrasting thermal properties of the underlying surface of zonally alternating oceans and continents and therefore can affect the wave. For $g > 0$ the system clearly shows chaotic behavior.

\begin{figure}[H]
    \centering
    \includegraphics[width=\textwidth]{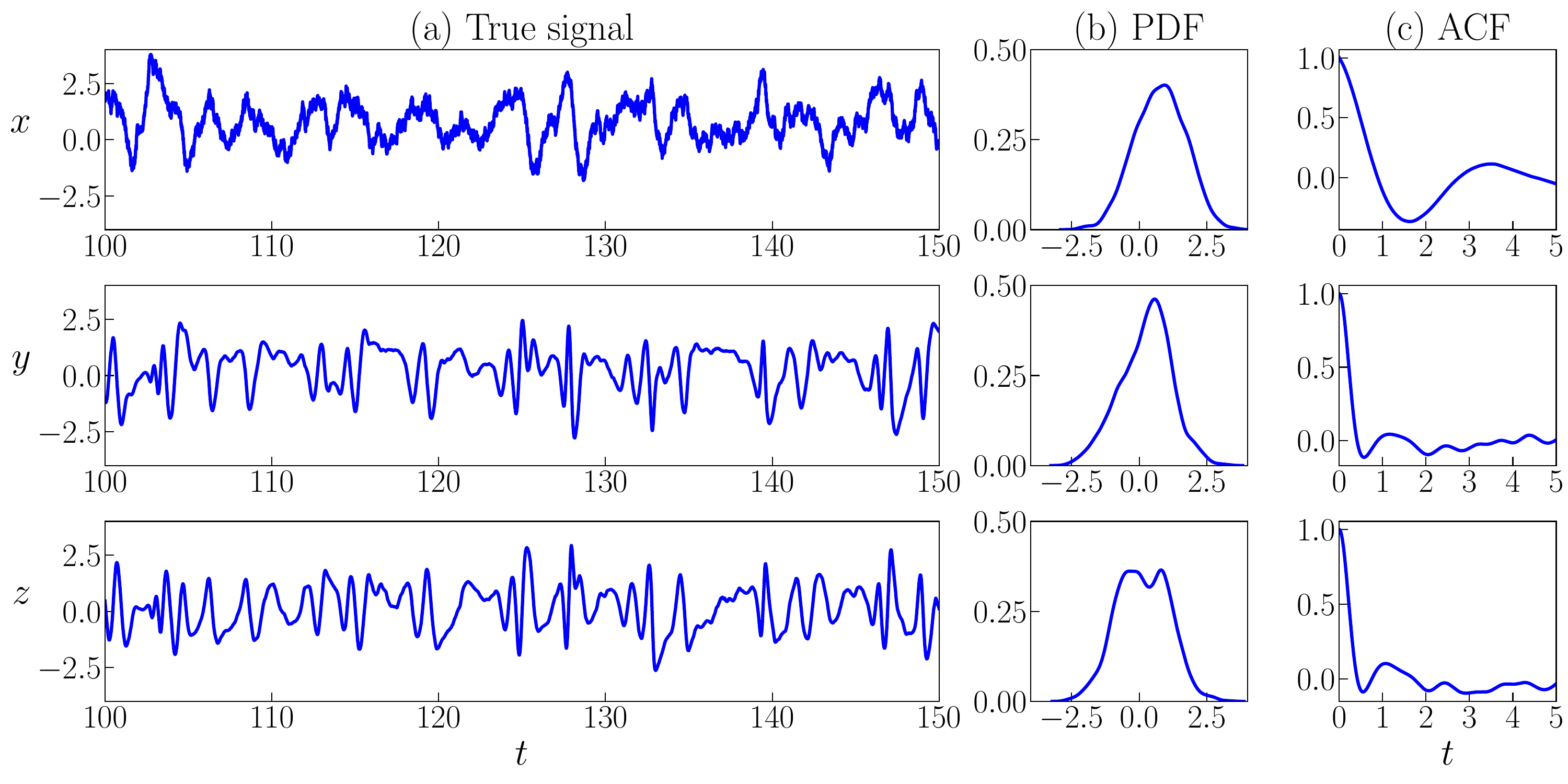}
    \caption{Model trajectories and the associated statistics of the true Lorenz 84 system. Panel (a): time series of each variable. Panel (b): the probability density function (PDF). Panel (c): the auto-correlation function (ACF). It should be noted that the PDFs and ACFs are estimated from much longer time series than the ones presented in Panel (a).}
    \label{fig:L84_Property}
\end{figure}

The parameter values used in the following tests are the standard values that create chaotic features:
\begin{equation}\label{L84_parameters}
  a = \frac{1}{4}, \quad b = 4, \quad f=8, \quad g=1, \quad \sigma_x = 1, \quad \sigma_y = 0.05, \quad\mbox{and}\quad \sigma_z = 0.05.
\end{equation}
See Figure \ref{fig:L84_Property} for a model simulation, the associated chaotic behavior, and the equilibrium statistics. Note that the decorrelation time (i.e., the integration of the autocorrelation function (ACF)) of the $x$ variable is about 0.5 time units, and those of the $y$ and $z$ variables are about 0.2 time units.

In the following test, $y$ and $z$ are assumed to be the observed variables while $x$ is unobserved. Note that a larger noise in the process of $x$ is taken to provide more variabilities and allow for the examination of the skill of state estimation. A time series of 250 units is generated, where the first 50 units are utilized for training, and the remaining 200 units are applied for testing.

As the Lorenz 84 is already a CGNS when $y$ and $z$ are treated as the observed variables, applying the causal inference approach in Section \ref{ssec:CGNSDE_Step1} will fully recover the exact dynamics. The additional neural network components are no longer needed in such a case. Since this numerical test example aims to illustrate the effectiveness of the CGNSDE with the DA loss, the physically explainable components in the hybrid model are manually specified, which can be regarded as using the user's prior knowledge to determine the explainable components of the model structure. Some terms in the original system are removed on purpose, which allows some room for the neural networks to improve the results. Note that the exact DA solution of the Lorenz 84 is available using \eqref{eq:CGNS_Filter}, which will serve as the reference solution to examine the performance of the CGNSDE.

With the prescribed knowledge-based components, the CGNSDE is given by:
\begin{equation}
    \label{eq:L84_CGNSDE}
    \begin{aligned}
        &\frac{\diff x}{\diff t} = f_x + a_xx + b_xz^2 + \mathbf{NN}_1(y,z;\theta) + \mathbf{NN}_4(y,z;\theta)x + \sigma_x\dot{W}_x,\\
        &\frac{\diff y}{\diff t} = f_y + a_yy  + \mathbf{NN}_2(y,z;\theta) + \mathbf{NN}_5(y,z;\theta)x + \sigma_y\dot{W}_y,\\
        &\frac{\diff z}{\diff t} = f_z + a_zz + b_zxz + \mathbf{NN}_3(y,z;\theta) + \mathbf{NN}_6(y,z;\theta)x + \sigma_z\dot{W}_z,
    \end{aligned}
\end{equation}
where $\mathbf{NN}_i$ is the $i$-th output of a single neural network $\mathbf{NN}: \mathbb{R}^2 \mapsto \mathbb{R}^6$ which is parameterized by $\theta$. The feed-forward neural network utilized here has 5 layers with 508 parameters. The training settings are $N_s = 200$ ($0.2$ units), $N_l = 50000$ ($50$ units) and $N_b=5000$ ($5$ units). The training starts with minimizing solely the forecast MSE \eqref{eq:forecast_mse} for the first 10000 epochs. Then the DA loss in \eqref{eq:DA_MSE} is included to minimize the total loss in \eqref{eq:total_MSE} for the subsequent 500 epochs. The optimizer for all models is selected as Adam with a learning rate $10^{-3}$.

Table \ref{tab:L84_Error} includes the performance of the models in a test period, which is different from the training data. The forecast MSE is based on 0.2 units prediction, while the DA MSE and DA negative log-likelihood are based on 200 units. It illustrates that the CGNSDE significantly outperforms the specified knowledge-based regression model regarding both state prediction and DA, indicating the critical role of the neural network components in the CGNSDE. The CGNSDE with the additional DA loss results in a smaller MSE in the posterior mean and a lower negative log-likelihood in the entire posterior distribution than the one with only the standard forecast MSE loss. This means the CGNSDE with the DA loss improves the DA skill with enhanced path-wise accuracy and reduced posterior uncertainty. Notably, the MSE and negative log-likelihood of the data assimilation solution using the true Lorenz 84 system are 0.0138 and -0.8175, respectively, similar to the result from the CGNSDE with the DA loss.

It is worth noticing that the CGNSDE without the DA loss is already quite skillful in DA. This is because the test model here is a CGNS. Therefore, even with only the forecast loss, the CGNSDE nearly captures the true dynamics, which leaves only a small room for the CGNSDE with the DA loss to improve the results further. In the following two numerical experiments, when the true system is not a CGNS, the advantage of the CGNSDE with the DA loss will become more significant, as the DA loss helps further improve the CGNSDE to characterize interdependence between different state variables.

\begin{table}[H]
    \centering
    \caption{Lorenz 84: Performance of three models in the test period including knowledge-based regression model, CGNSDE without DA loss, and CGNSDE with the DA loss. }
    \label{tab:L84_Error}
    \begin{tabular}{|c|c|c|c|}
        \hline
        &Forecast MSE & DA MSE & DA Neg-Log-Likelihood \\
        \hline
        Knowledge-based regression model & 0.2114 & 8.5685 & 20.8234 \\
         \hline
        CGNSDE without DA loss & 0.0487 & 0.0444 & 0.6289  \\
        \hline
        CGNSDE with the DA loss & 0.0469 & 0.0183 & -0.5870  \\
        \hline
    \end{tabular}
\end{table}

\begin{figure}[H]
    \centering
    \includegraphics[width=\textwidth]{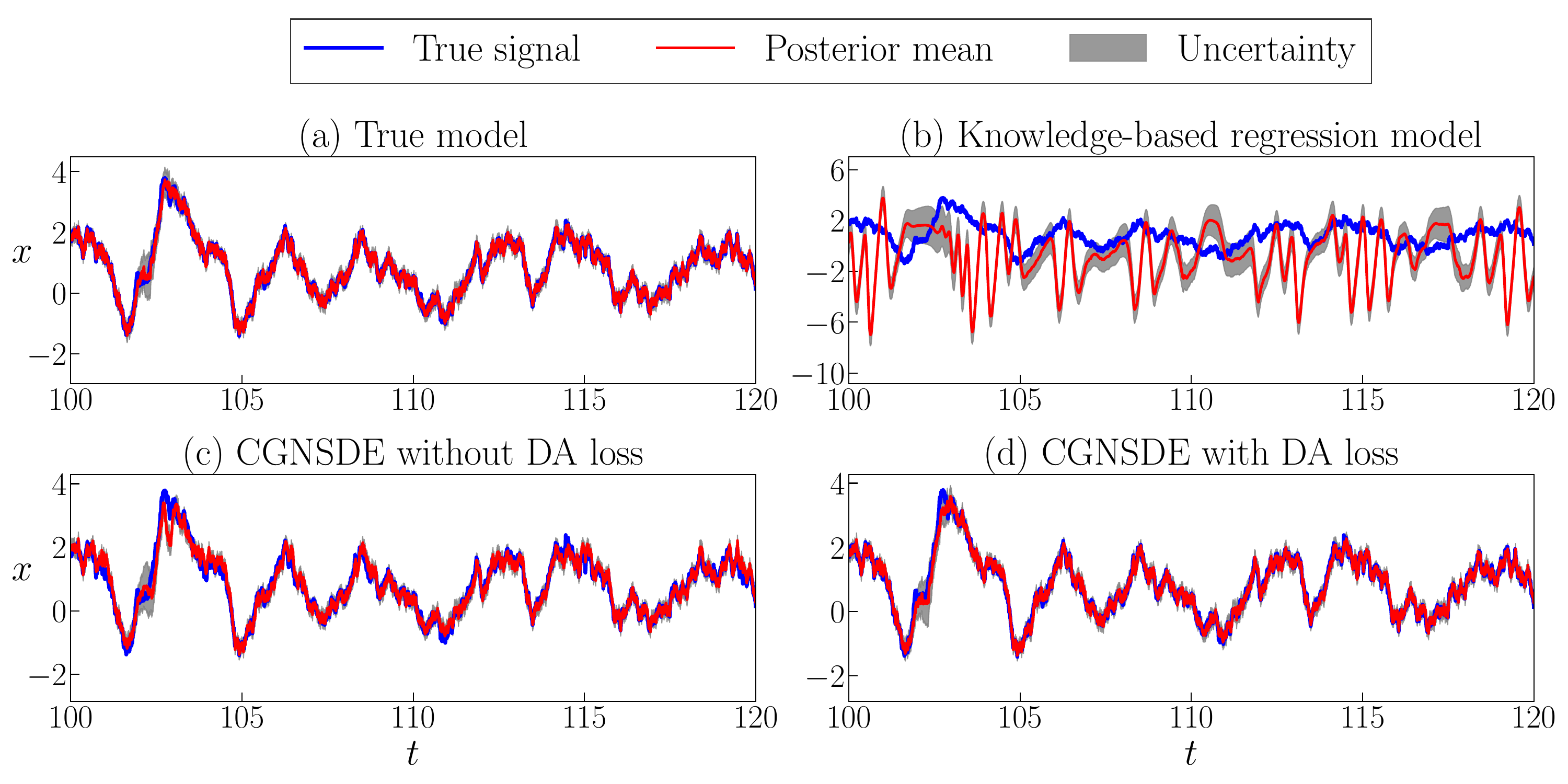}
    \caption{DA results of Lorenz 84 system from the closed analytic formulae in \eqref{eq:CGNS_Filter} for true system and three models. The uncertainties are indicated by the grey colored regions, which correspond to two standard deviation from the posterior mean.}
    \label{fig:L84_DA}
\end{figure}

The findings in Table \ref{tab:L84_Error} are validated by Figure \ref{fig:L84_DA}, which shows the DA results using the true system and three models as the forecast model, respectively. The two CGNSDEs (with and without DA loss) significantly improve the performance of the DA compared with the pre-determined knowledge-based regression models, indicating the necessity of incorporating the neural network components in the CGNSDE. The CGNSDE with the DA loss can further improve the DA accuracy compared to its counterpart with only the standard forecast MSE as the loss function.

\begin{figure}[H]
    \centering
    \includegraphics[width=\textwidth]{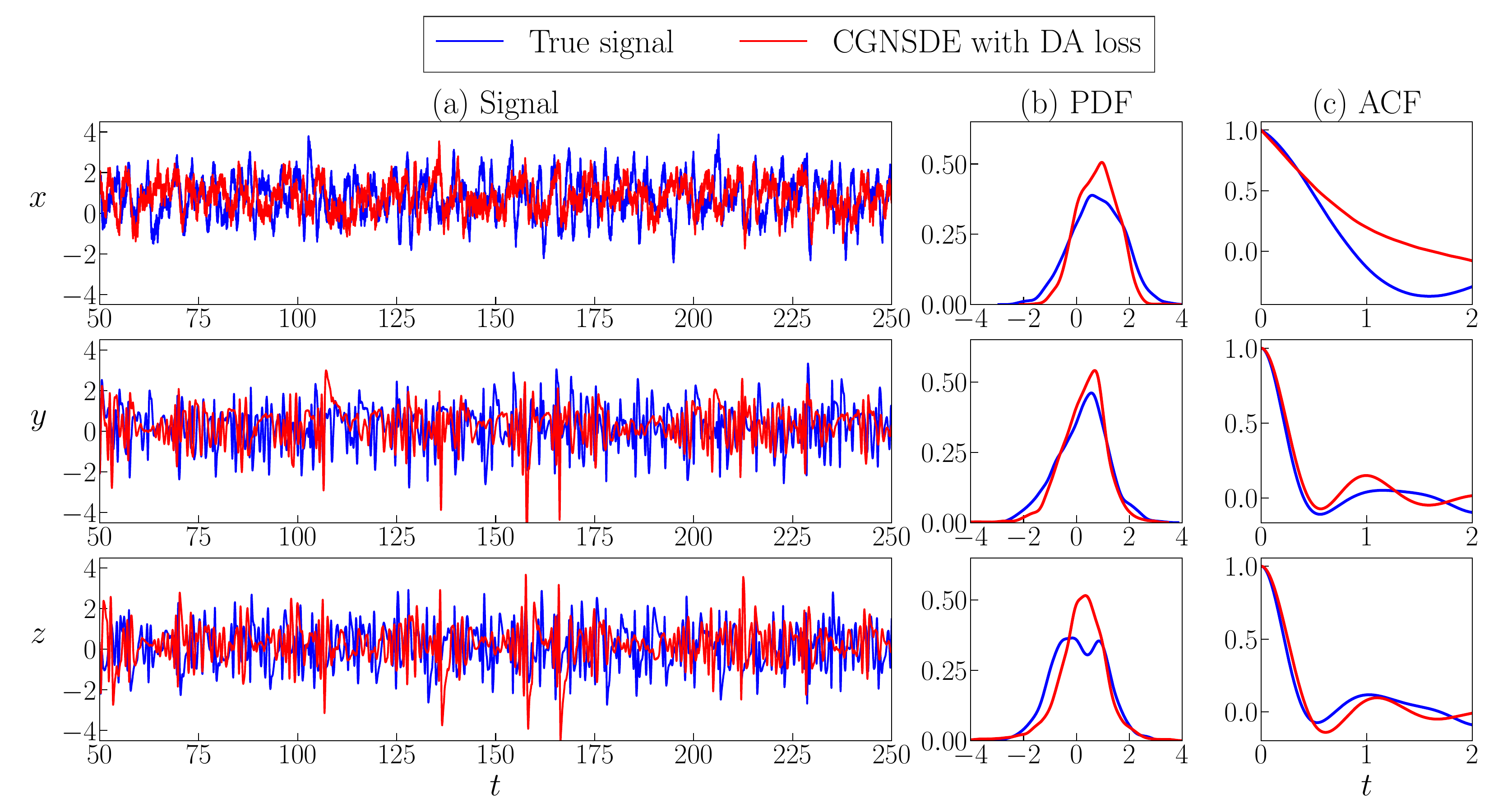}
    \caption{One long-term realization of the CGNSDE model trained with both forecast and DA losses for the Lorenz 84 system. Panel (a): time series of different variables. Panel (b) and (c): the associated PDFs and ACFs.}
    \label{fig:L84_LongSimu}
\end{figure}

Finally, Figure \ref{fig:L84_LongSimu} shows one long-term simulation of the CGNSDE with the DA loss. Panel (a) indicates that the CGNSDE can provide stable long-term simulations without explicitly regularizing its stability in training. The overall patterns of the simulated time series are similar to the true system results. Panels (b) and (c) show that the long-term statistics (e.g., the PDFs and the ACFs) associated with the simulations from the CGNSDE have a qualitative agreement with the true system, despite utilizing only short time series as the forecast loss in the training stage.

\subsection{The projected stochastic Burgers–Sivashinsky equation: a highly nonlinear system with intermittency}
\label{ssec:results_BSE}

The Fourier-Galerkin projection of stochastic Burgers-Sivashinshky equation is a three-dimensional SDEs with energy-conserving quadratic nonlinear terms and subject to additive white noise forcing  \citep{chekroun2014stochastic, chekroun2016post}:

\begin{equation}
    \begin{aligned}\label{eq:PSBSE}
        \frac{\diff x}{\diff t} &= \beta_xx + \alpha xy + \alpha yz + \sigma_x\dot{W}_x,\\
        \frac{\diff y}{\diff t} &= \beta_yy - \alpha x^2 + 2\alpha xz + \sigma_y\dot{W}_y,\\
        \frac{\diff z}{\diff t} &= \beta_zz - 3\alpha xy + \sigma_z\dot{W}_z,
    \end{aligned}
\end{equation}
where the coefficients for the linear terms are chosen such that $\beta_x$ is positive to introduce linear instability into the system, while $\beta_y$ and $\beta_z$ are negative, representing linear damping effects. The coefficient $\alpha>0$ controls the strength of the nonlinearity. The noise strength coefficients $\sigma_x$, $\sigma_y$, and $\sigma_z$ are positive constants. This system can, for instance, be obtained as a Fourier-Galerkin projection of the stochastic Burgers-Sivashinsky equation
\begin{equation*}
\frac{\partial u}{\partial t}  = \big( \nu \partial_{xx} u  + \lambda u  - u  \partial_x u\big) + \dot{W}(t,x)
\end{equation*}
posed on a bounded interval $x \in (0,L)$ subject to homogeneous Dirichlet boundary conditions. In this context, $\beta_x$, $\beta_y$, and $\beta_z$ are simply the three largest eigenvalues of the linear operator, and $\alpha$ is linked to the domain size $L$ via $\alpha = \pi/(\sqrt{2} L^{3/2})$.

In the following, the largest-scale variable $x$ is treated as the observed variable while there are no direct observations for $y$ and $z$. Under this splitting of the state variables, system \eqref{eq:PSBSE} does not have the conditional Gaussian structure due to the quadratic nonlinear term $\alpha yz$ between the unobserved variables that appear in \eqref{eq:PSBSE}. Nevertheless, due to the low dimensionality, the ensemble Kalman-Bucy filter (EnKBF) \citep{bergemann2012ensemble} can be utilized to provide a reference solution of DA.

\begin{figure}[H]
    \centering
    \includegraphics[width=\textwidth]{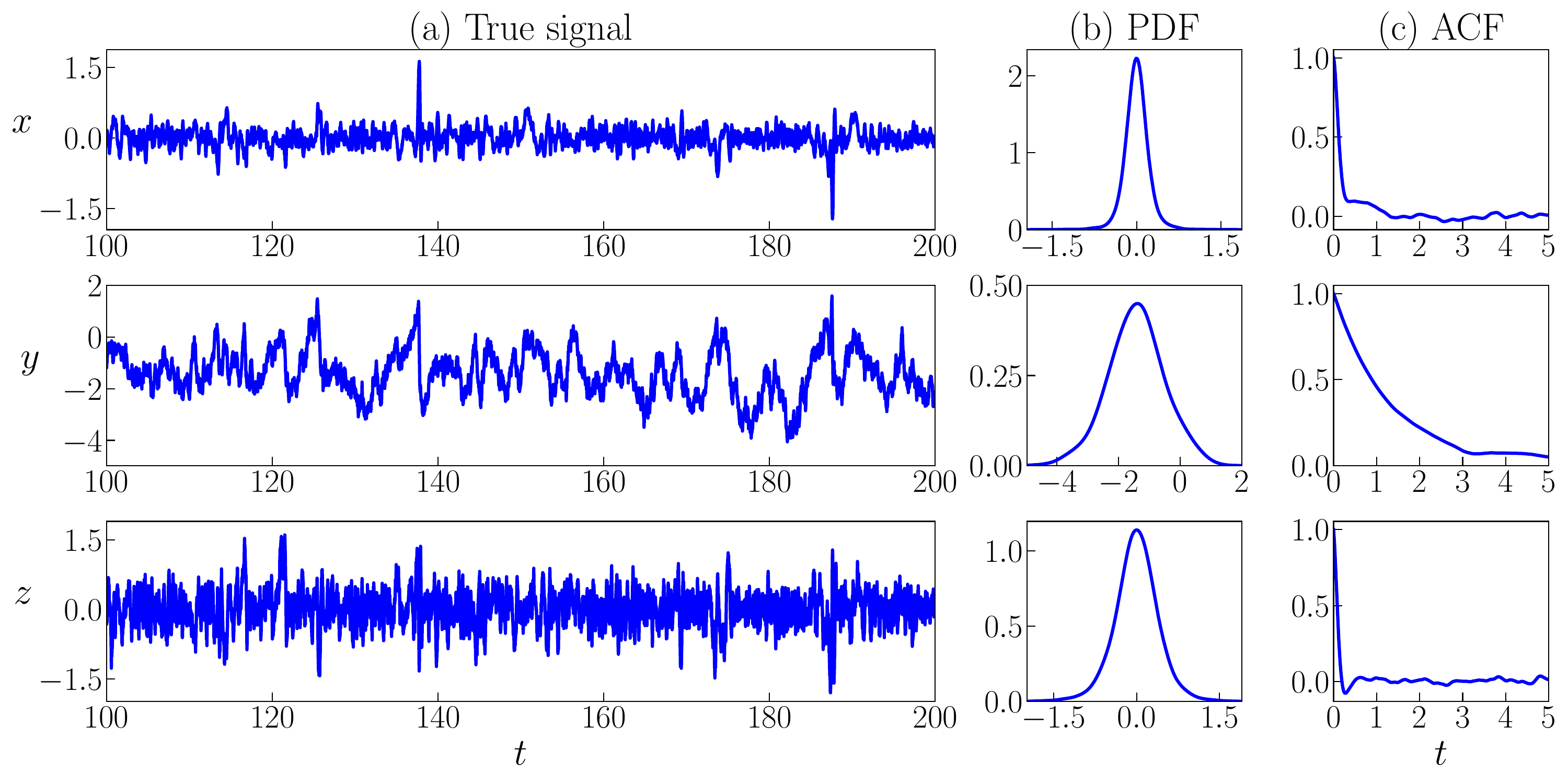}
    \caption{One simulation of the projected stochastic Burgers–Sivashinsky equation \eqref{eq:PSBSE}. Panel (a): time series of each state variable. Panel (b): the PDFs. Panel (c): the ACFs. It should be noted that the PDFs and ACFs are estimated from much longer simulations than the one presented in Panel (a).}
    \label{fig:PSBSE_Property}
\end{figure}

The parameter values used in the following tests are the standard values that create strongly intermittent features with extreme events and highly non-Gaussian PDFs:
\begin{equation}
  \beta_x=0.2, \quad \beta_y=-0.3, \quad\beta_z=-0.5,\quad \alpha = 5,\quad \sigma_x=0.3, \quad\sigma_y=1, \quad\mbox{and}\quad\sigma_z = 1.
\end{equation}
A time series of 600 units is generated, where the first 100 units are utilized for training, and the remaining 500 units are applied for testing. Figure \ref{fig:PSBSE_Property} shows a period of the model simulation and the associated statistics. The time series display strong intermittent behavior with non-Gaussian statistics and extreme events.

To develop a CGNSDE, the causal inference method described in Section \ref{ssec:CGNSDE_Step1} is utilized to identify the knowledge-based components of system dynamics. The library is constructed with candidate functions containing all possible linear and quadratic nonlinear functions that allow the model to satisfy the conditional Gaussian structures. Therefore, the library is given by $\{x, y, z, x^2, xy, xz\}$. The values of the resulting causation entropies from these candidate functions to the target dynamics are included in Table \ref{tab:PSBSE_CEM}. It is worth noting that the relative strengths of the causation entropies should be compared only within each row. Different rows, presenting the governing equations of different state variables, are independent of the others. The causation entropies in different rows may have significant differences in the order of amplitudes. By considering the relative magnitudes within each row, the candidate functions ($z$, $xy$) are selected for the $x$ dynamics, ($x^2$, $xz$) are for the $y$ dynamics, and $xy$ is for the $z$ dynamics.



\begin{table}[H]
    \centering
    \caption{The projected stochastic Burgers–Sivashinsky equation: causation entropy between dynamics and candidate functions in the library. The significant values, corresponding to the terms used to build the knowledge-based components in the CGNSDE, are highlighted in bold font. Note that the relative strengths of the causation entropies should be compared only within each row.}
    \label{tab:PSBSE_CEM}
    \begin{tabular}{|c|c|c|c|c|c|c|}
        \hline
         &$x$ &$y$ &$z$ &$x^2$ &$xy$ &$xz$ \\
         \hline
         $\dot{x}$ & 0.000 & 0.000 & \textbf{0.093} & 0.000 &\textbf{0.049} & 0.001 \\
         \hline
         $\dot{y}$ & 0.000 & 0.000 & 0.000 & \textbf{0.004} & 0.000 & \textbf{0.011} \\
         \hline
         $\dot{z}$ & 0.000& 0.000 & 0.001&  0.000 & \textbf{0.070} & 0.000      \\
         \hline
    \end{tabular}
\end{table}

After supplementing neural networks to the knowledge-based regression model, the CGNSDE for this projected stochastic Burgers–Sivashinsky equation reads:
\begin{equation}
    \label{eq:PSBSE_CGNSDE}
    \begin{aligned}
        \frac{\diff x}{\diff t} &= a_xz + c_xxy + \mathbf{NN}_1(x;\theta) + \mathbf{NN}_4(x;\theta)y + \mathbf{NN}_5(x;\theta)z + \sigma_x\dot{W}_x\\
        \frac{\diff y}{\diff t} &= b_yx^2 + c_yxz + \mathbf{NN}_2(x;\theta)+\mathbf{NN}_6(x;\theta)y + \mathbf{NN}_7(x;\theta)z +\sigma_y\dot{W}_y\\
        \frac{\diff z}{\diff t} &= c_zxy + \mathbf{NN}_3(x;\theta)+ \mathbf{NN}_8(x;\theta)y + \mathbf{NN}_9(x;\theta)z + \sigma_z\dot{W}_z
    \end{aligned}
\end{equation}
In the CGNSDE \eqref{eq:PSBSE_CGNSDE}, $\mathbf{NN}_i$ is the $i$-th output of a single neural network $\mathbf{NN}: \mathbb{R}^1 \mapsto \mathbb{R}^9$ which is parameterized by $\theta$. The feed-forward neural network utilized here has 5 layers with 449 parameters.

The setup for training the two CGNSDEs is: $N_s=50$ ($0.5$ units), $N_l=10000$ ($100$ units), $N_b=1000$ ($10$ units). The CGNSDE without DA is trained for 10000 epochs to minimize the forecast MSE \eqref{eq:forecast_mse}. The CGNSDE with DA is achieved by retraining CGNSDE for another 500 epochs to minimize total loss \eqref{eq:total_MSE} including DA loss. The optimizer for all models is selected as Adam with a learning rate of $10^{-3}$.

The test results of all three models are summarized in Table \ref{tab:PSBSE_Error}. In the test period, the forecast MSE is calculated by state prediction with 0.5 units, while the DA MSE and DA negative log-likelihood are based on 500 units. Although the CGNSDE trained without the DA loss gives the lowest error for short-term prediction, which is around the level of the intrinsic error due to the random noise in the true system, it has the largest DA MSE. The error is even more significant than the knowledge-based regression model. In contrast, the CGNSDE model trained with both the forecast and the DA losses in \eqref{eq:total_MSE} can significantly enhance the DA performance with a slight trade-off of short-term forecast accuracy. This improvement indicates the essential role of including the DA loss in training the CGNSDE.

It should be noted that the true system is a non-CGNS in this example, and thus, the DA using the true system cannot be done by applying the explicit formulae in \eqref{eq:CGNS_Filter}. Therefore, the EnKBF is adopted to obtain the DA solution of the true system as a reference. The number of ensemble members in the EnKBF is set to be $J=100$. Using the original noise parameters ($\sigma_x = 0.3, \sigma_y = 1, \sigma_z=1$), the EnKBF will encounter a catastrophic filter divergence \citep{kelly2015concrete}, i.e., the solution has a finite-time numerical blow-up issue. By applying a standard noise inflation strategy \citep{asch2016data}, the minimum DA MSE and the corresponding negative log-likelihood of applying the EnKBF to the true system are 0.2237 and 7.3375, respectively. Notably, the DA MSE of the CGNSDE with the DA loss is 0.2674, which is comparable with the solution given by EnKBF. Since the CGNSDE framework does not demand ensemble simulations, it can reduce the computational cost and avoid potential sampling errors due to an insufficient ensemble size.

\begin{table}[H]
    \centering
    \caption{The projected stochastic Burgers–Sivashinsky equation: Performance in the test period of the knowledge-based regression model, the CGNSDE without DA loss, and the CGNSDE with the DA loss.}
    \label{tab:PSBSE_Error}
    \begin{tabular}{|c|c|c|c|}
        \hline
        &Forecast MSE & DA MSE & DA Neg-Log-Likelihood \\
        \hline
        Knowledge-based regression model & 0.1799 & 0.8589 & 2.9459  \\
         \hline
        CGNSDE without DA loss & 0.1347 & 1.2907 & 4.4199 \\
        \hline
        CGNSDE with the DA loss & 0.1446 & 0.2674 & 1.3528\\
        \hline
    \end{tabular}
\end{table}

\begin{figure}[H]
    \centering
    \includegraphics[width=\textwidth]{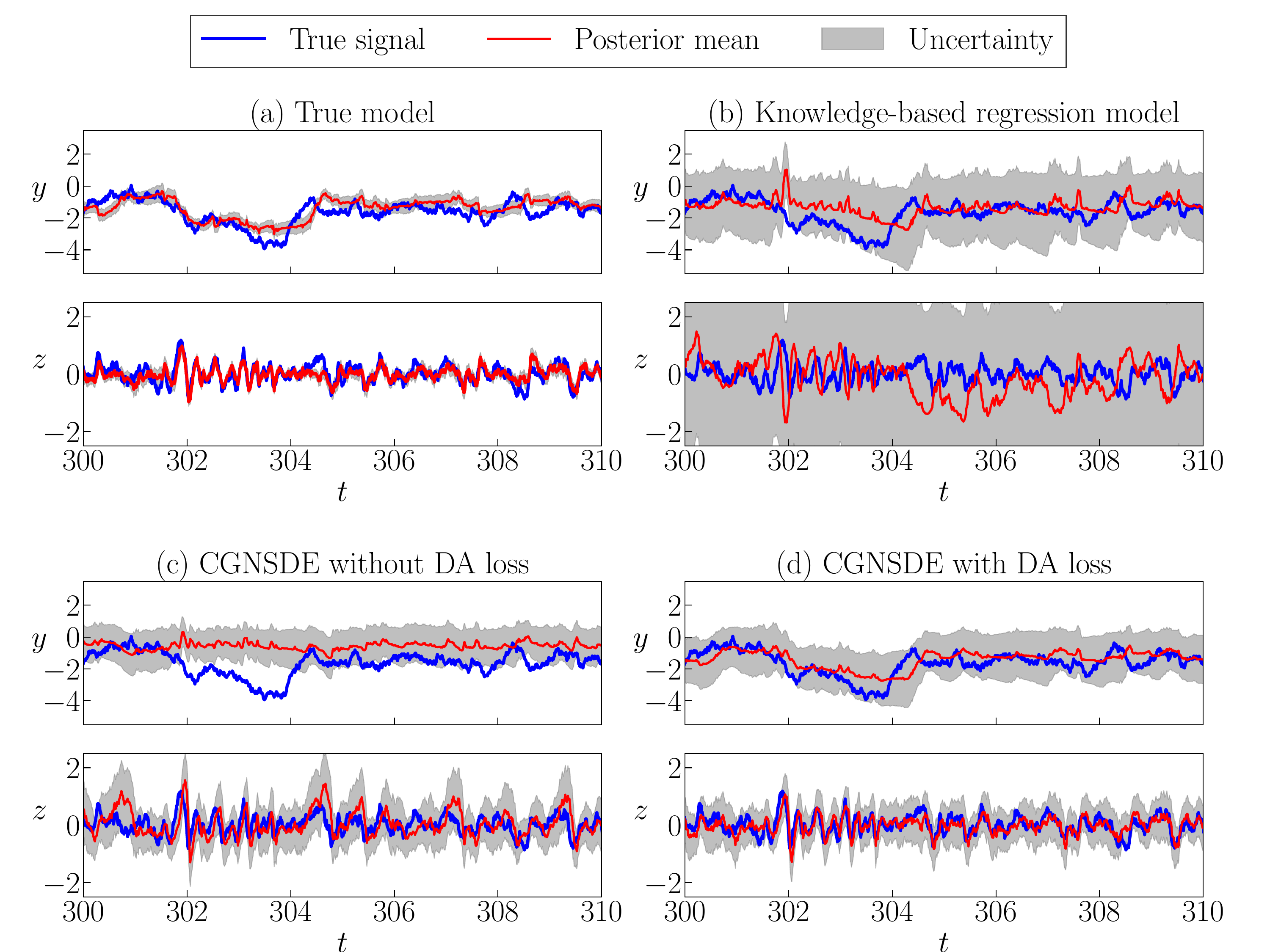}
    \caption{DA results of the projected stochastic Burgers–Sivashinsky equation from EnKBF for true system and closed analytic formulae in \eqref{eq:CGNS_Filter} for three models. The uncertainties are indicated by the grey colored regions, which correspond to two standard deviations from the posterior mean.}
    \label{fig:PSBSE_DA}
\end{figure}

Figure \ref{fig:PSBSE_DA} compares the DA results by applying the EnKBF to the true system and by applying the closed analytic formulae to the regression model and the two CGNSDEs. Panel (a) indicates that the results of EnKBF for the true system still have a noticeable difference from the true states $y$ and $z$. Despite tuning the noise coefficients to minimize the error in the posterior mean, the posterior uncertainty seems severely underestimated. In contrast, the posterior mean estimates from the CGNSDE trained with the DA loss, as shown in Panel (d), demonstrate a similar performance as the EnKBF for the true system, while the range of the associated uncertainty of the CGNSDE with the DA loss can cover the truth, including the extreme events. The results confirm the advantages of the CGNSDE framework with DA for systems with highly non-Gaussian features. On the other hand, both the regression model and the CGNSDE without the DA loss lead to much larger biases in the DA solutions.

\begin{figure}[H]
    \centering
    \includegraphics[width=\textwidth]{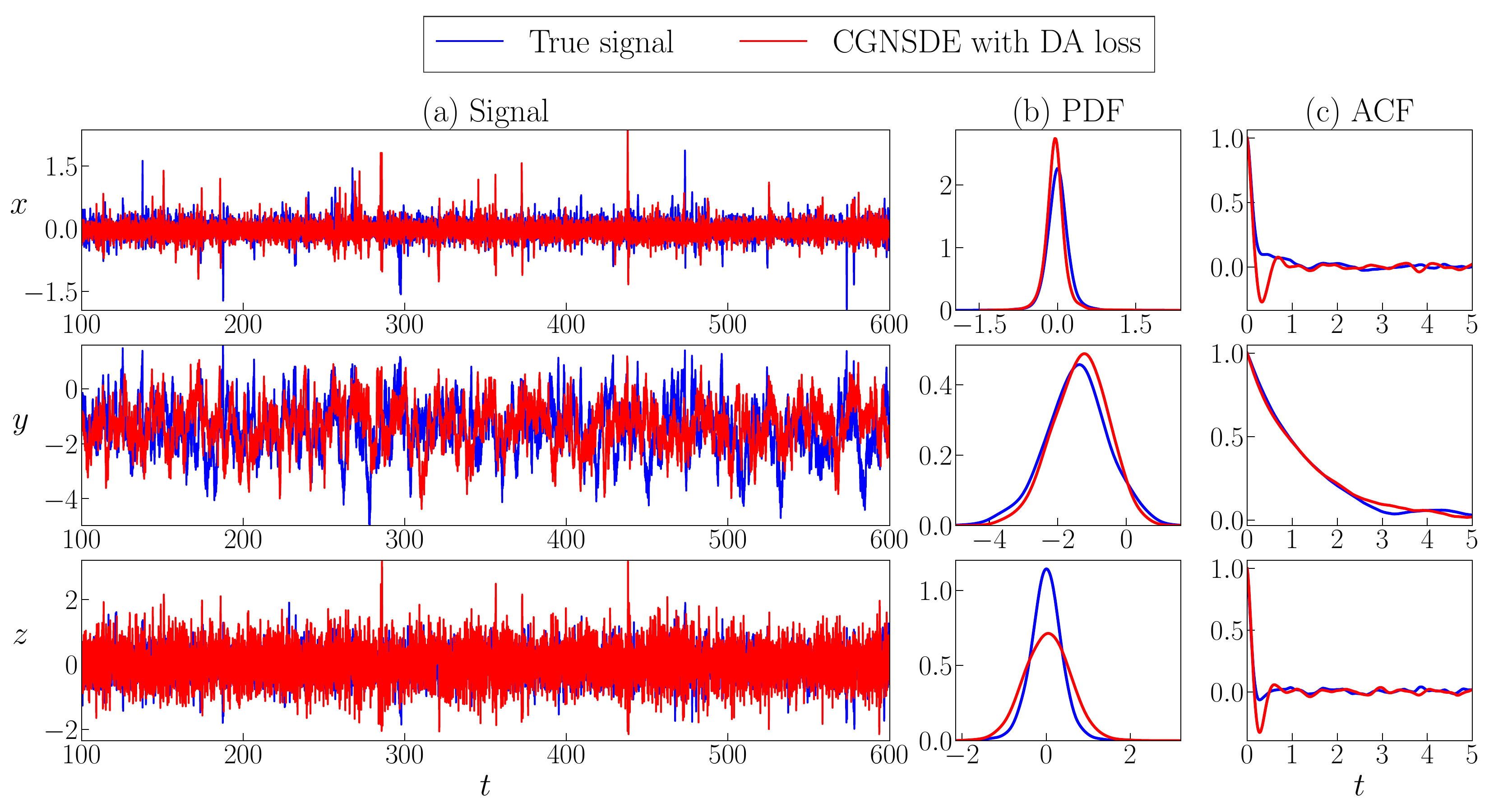}
    \caption{Long-term simulations of the CGNSDE with the DA loss starting from the initial state of test data for the projected stochastic Burgers–Sivashinsky equation. Panel (a): time series of each state. Panels (b) and (c): PDFs and ACFs.}
    \label{fig:PSBSE_LongSimu}
\end{figure}

Figure \ref{fig:PSBSE_LongSimu} compares the long-term simulation results of the CGNSDE with the DA loss to the true signal. Similar to the previous example, the CGNSDE can also provide effective long-term simulations, demonstrating a qualitative agreement with the time series of the true system as presented in Panel (a). Remarkably, the long-term time series of the modeled data can reproduce the overall pattern as the true signal with noticeable extreme events in the $x$ and $z$ processes. Panels (b) and (c) show the associated PDFs and the ACFs. The PDFs of $x$ and $y$ from the CGNSDE are nearly identical to the truth, especially in capturing the non-Gaussian features. The PDF of $z$ has a slightly heavier tail. Note that the model is trained based only on short data. The skillful long-term forecast result is partially due to the help from the additional DA loss, which highlights the dynamical coupling between different state variables that advances the long-term behavior of the system. Similarly, the ACFs from the CGNSDE are nearly the same as the truth. In contrast, the long-term simulation from the knowledge-based model has highly oscillated patterns in the $y$ and $z$ dynamics, while the one from CGNSDE without DA loss cannot reproduce the extreme events in $x$ dynamics (not shown here).

\subsection{The Lorenz 96 model: A moderate-dimensional chaotic system}
\label{ssec:results_L96}

The Lorenz 96  model with noise is given by \citep{lorenz1996predictability, ott2002exploiting}
\begin{equation}
    \begin{aligned}\label{eq:Lorenz96}
        \frac{\diff x_i}{\diff t}=(x_{i+1} - x_{i-2})x_{i-1} - c_ix_i +F_i + \sigma_i\dot{W}_i, \quad i=1,2,3 \cdots, I.
    \end{aligned}
\end{equation}
The model can be regarded as a coarse discretization of atmospheric flow on a latitude circle with complicated wave-like and chaotic behavior. It schematically describes the interaction between small-scale fluctuations with larger-scale motions. It is widely used as a testbed for DA, state forecast, uncertainty quantification, and parameterization in numerical weather forecasting \citep{wilks2005effects, arnold2013stochastic}. Notably, depending on the choice of the set of the observed and unobserved variables, the Lorenz 96 equation \eqref{eq:Lorenz96} can be a CGNS or a non-CGNS. The study based on the Lorenz 96 model includes three cases:
\begin{enumerate}
    \item The true dynamics is a CGNS with spatially homogeneous statistics. The system has constant parameters such that the spatial patterns are statistically homogeneous. In other words, different grids, indicated by different indices $i$ in \eqref{eq:Lorenz96}, have the same equilibrium statistics. Two-thirds of the states are observable, and the other one-third are unobserved. In such a way, the true model becomes a CGNS. The primary purpose of this test is to demonstrate the capability of the CGNSDE model in handling relatively high-dimensional systems.
    \item The true dynamics is a non-CGNS with spatially homogeneous statistics. The same spatially homogeneous system is utilized as in Case 1. Half of the states are observable, and the other half are unobserved, so the true system becomes a non-CGNS. The focus of this case is on demonstrating the capability of the CGNSDE in handling a non-CGNS with a relatively high dimension.
    \item The true dynamics is a non-CGNS with spatially inhomogeneous statistics. The system has spatially dependent parameters, so the statistical behavior at different grid points is inhomogeneous. The true system is a non-CGNS, with half of the states being observable and the other half being unobservable. The main focus of this case is on demonstrating that the performance of using the CGNSDE with a simple translate-invariant neural network structure can handle spatially inhomogeneous dynamics.
\end{enumerate}

The following parameter values are adopted as in the true system for the first two cases with homogeneous dynamics:
\begin{equation}\label{L96_parameters}
I=36, \qquad F_i=8, \qquad c_i = 1, \qquad\mbox{and}\qquad\sigma_i=0.5,
\end{equation}
and the parameters for the third case with inhomogeneous statistics are as follows,
\begin{equation}\label{L96_parameters2}
I=36, \qquad F_i=8, \qquad c_i = 2+1.5\sin(2\pi (i-1)/I), \qquad\mbox{and}\qquad\sigma_i=0.5.
\end{equation}
A time series of 300 units is generated, where the first 100 units are utilized for training, and the remaining 200 units are applied for testing.

It is worth highlighting that the nonlinearity in Lorenz 96 is given by advection. Due to its localized interactions between state variables, the complexity of the neural network part can be significantly reduced. In particular, the number of neural networks will not increase as a function of the dimension of the underlying system.

\subsubsection{Case 1: CGNS that is spatially homogeneous}
\label{sssec:results_L96_1}
\begin{figure}[H]
    \centering
    \includegraphics[width=0.7\textwidth]{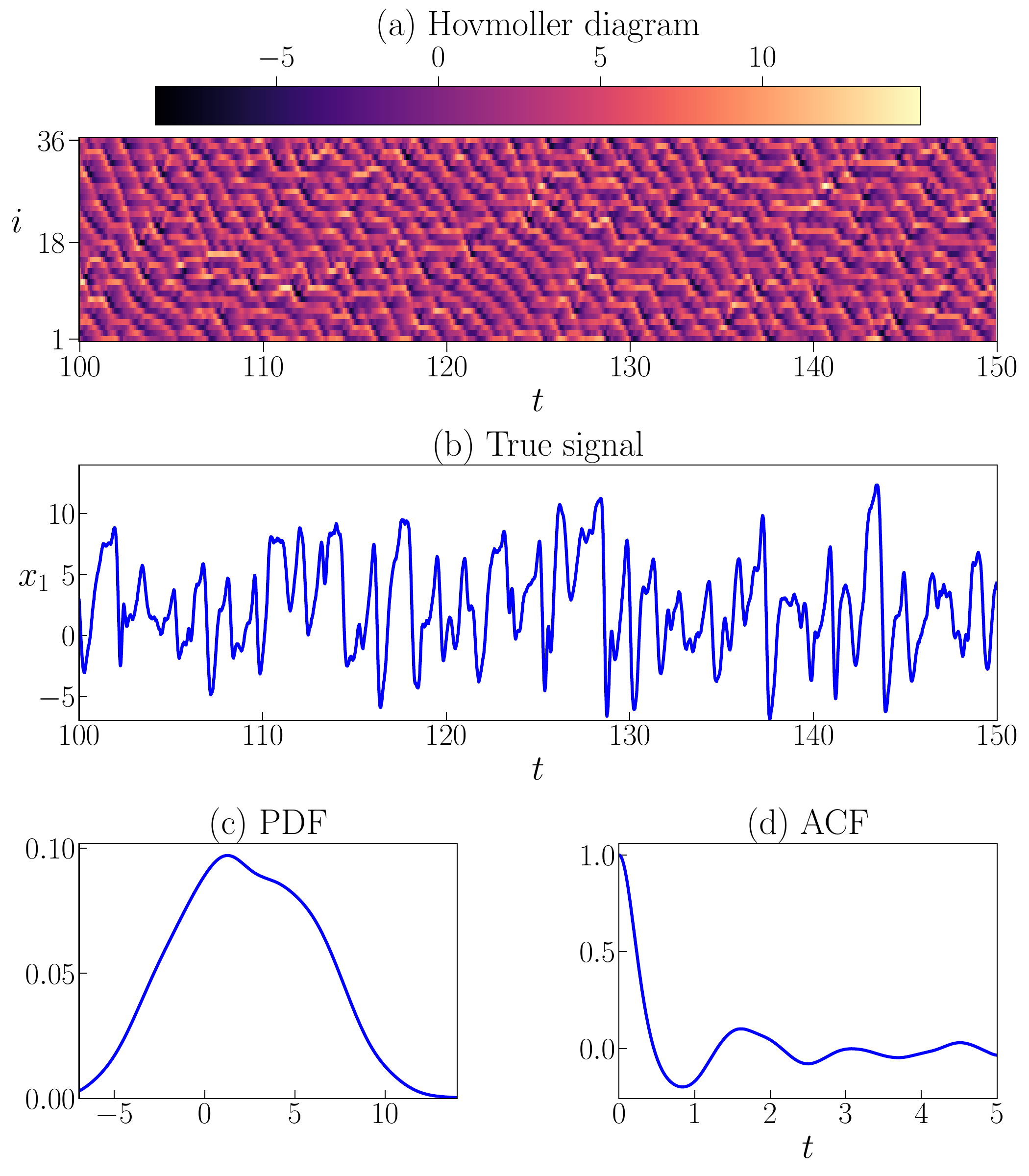}
    \caption{Model simulation and statistics of the noisy Lorenz 96 system with parameters in \eqref{L96_parameters}. Panel (a): the Hovmoller diagram of the spatiotemporal patterns. Panel (b): time series of $x_1$. Panel (c): the PDF of $x_1$. Panel (d): the ACF of $x_1$. Note that the PDF and ACF are estimated from a simulation longer than the one presented in Panel (b). The behavior at other grid points is similar since the system is statistically homogeneous in space.}
    \label{fig:L96_Property}
\end{figure}

Figure \ref{fig:L96_Property} includes a model simulation and the associated statistics of the true Lorenz 96 system \eqref{eq:Lorenz96} with parameters in \eqref{L96_parameters}. Panel (a) shows the spatiotemporal patterns (i.e., the Hovmoller diagram), where the propagation of waves and the chaotic pattern of the system can be seen. Panel (b) shows the time series of $x_1$, where the associated PDFs and ACFs are displayed in Panels (c)--(d). The behavior at other grid points is similar since the system is statistically homogeneous in space.

In this case, the observed state variables are $\mathbf{u}_1 = [x_1, x_2, x_4, x_5, \cdots, x_{34}, x_{35}] \in \mathbb{R}^{24} $ and the unobserved variables are $\mathbf{u}_2 = [x_3, x_6, \cdots, x_{36}] \in \mathbb{R}^{12}$. With such a choice, the Lorenz 96 system can be formalized as a CGNS in \eqref{eq:CGNS}.

Since the Lorenz 96 model in such a case is already a CGNS, if the causal inference is applied, then the knowledge-based model will fully recover the true dynamics. As in the Lorenz 84 test case, an intrinsic bias in the prior knowledge is introduced on purpose to allow neural networks to play some roles in improving the model behavior. To this end, the candidate function library is assumed only to contain linear functions. Therefore, applying causal inference leads to the part of the knowledge-based regression model with the correct linear structure of the original system. The remaining information of the system will be characterized by neural networks.

With the linear terms identified by the causal inference, the CGNSDE reads
\begin{equation}
\label{eq:L96(case1)_CGNSDE}
\frac{\diff x_i}{\diff t} =
\begin{cases}
    \text{if } i \bmod 3 = 1: \\
    \quad \begin{aligned}
        \Tilde{F} + \Tilde{c}x_i + &\mathbf{NN}^{(1)}_1(x_{i-2}, x_i, x_{i+1};\theta_1) + \\
        &\mathbf{NN}^{(1)}_2(x_{i-2}, x_i, x_{i+1};\theta_1)x_{i-1}+\\
        &\mathbf{NN}^{(1)}_3(x_{i-2}, x_i, x_{i+1};\theta_1)x_{i+2}+ \sigma_i\dot{W}_i,
    \end{aligned} \\
    \text{if } i \bmod 3 = 2: \\
    \quad \begin{aligned}
        \Tilde{F} + \Tilde{c}x_i + &\mathbf{NN}^{(2)}_1(x_{i-1}, x_{i}, x_{i+2};\theta_2) + \\
        &\mathbf{NN}^{(2)}_2(x_{i-1}, x_{i}, x_{i+2};\theta_2)x_{i-2} +\\
        &\mathbf{NN}^{(2)}_3(x_{i-1}, x_{i}, x_{i+2};\theta_2)x_{i+1}+ \sigma_i\dot{W}_i,
    \end{aligned} \\
    \text{if } i \bmod 3 = 0: \\
    \quad \begin{aligned}
        \Tilde{F} + \Tilde{c}x_i + &\mathbf{NN}^{(3)}_1(x_{i-2}, x_{i-1}, x_{i+1}, x_{i+2};\theta_3) +\\
        &\mathbf{NN}^{(3)}_2(x_{i-2}, x_{i-1}, x_{i+1}, x_{i+2};\theta_3)x_i+ \sigma_i\dot{W}_i,
    \end{aligned}
\end{cases}
\end{equation}
where ``$\bmod$'' denotes the standard modulo operation that calculates the remainder, and $i=1, 2,\cdots, 36$. The $\mathbf{NN}^{(i)}_j$ in \eqref{eq:L96(case1)_CGNSDE} stands for the $j$-th output of $i$-th neural network. The CGNSDE is developed by taking advantage of the fact that state variables only have localized dependence. For $i \bmod 3 =1$, $x_i$ are the observed variables, and their nearby dependent state variables are $(x_{i-2},x_{i-1},x_{i+1},x_{i+2})$, in which $x_{i-1}$ and $x_{i+2}$ are unobserved. Therefore, the first part of \eqref{eq:L96(case1)_CGNSDE} is constructed to satisfy the CGNS form, with linear dependency on nearby hidden states $x_{i-1}$ and $x_{i+2}$. The second part of \eqref{eq:L96(case1)_CGNSDE} can be constructed similarly, while the third part has a different form from the previous two, mainly because $x_i$ are unobserved state variables for $i \bmod 3=0$. The three neural networks all have 3 layers with 57, 57, and 64 parameters, respectively.

It is worth noting that the same neural network (with the same structures and parameters) is utilized for each set of state variables with different indices $i$. For example, the first neural network $\mathbf{NN}^{(1)}$ appears in the governing equations of $x_1, x_4,\ldots, x_{34}$. This is due to the translate-invariant structure of the Lorenz 96 system, which is assumed to be a known feature in developing the CGNSDE. By applying such a physical property, the complexity of the neural network part can be significantly reduced since, otherwise, the number of neural networks will scale as a function of the dimension of the underlying system. Furthermore, when the statistics are homogeneous, the effective length of the training data equals the actual length of the observations multiplied by the number of grids. This means that with a large number of grids, a short segment of the model observations is usually sufficient to train the neural network.

The CGNSDE without the DA loss has been trained by minimizing the forecast MSE in \eqref{eq:forecast_mse} with $N_{s}=1$. This training method based on one-step state prediction is formalized by training the model to approximate the mapping from states to dynamics in practice. Then the CGNSDE with the DA loss will be retrained by $N_{s}=5$, $N_l=10000$ and $N_b=500$, which means the matching time of states prediction is $0.05$ time units and the DA generation time is $100$ time units with the first $5$ time units being omitted as the burn-in period for DA. The Adam optimizer is selected with $10^{-3}$ learning rate.

The performance of the three models in the test period is summarized in Table \ref{tab:L96(case1)Error}. The forecast MSE is calculated by 0.2 units of state prediction, while the DA MSE and DA negative log-likelihood are based on 200 units. The knowledge-based model with only simple linear structures fails to approximate the underlying dynamics from data, resulting in substantial errors in both forecast and DA. Compared to this knowledge-based regression model, the CGNSDE without DA loss can significantly improve the forecast and DA performance after introducing the neural network components. By further incorporating DA loss in training, the CGNSDE slightly reduces the DA error in the test period but somewhat increases the forecast error. Notably, the MSE and negative log-likelihood of the data assimilation solution using the true Lorenz 96 system are 0.05284 and 1.0723, respectively, similar to the results from the two CGNSDEs. The comparable performance of the two CGNSDEs is not too surprising, since the true system is a CGNS so the CGNSDE with only the forecast loss may have already fully captured the true underlying dynamics. However, as seen below, this will not be the case when the true system is a non-CGNS.

\begin{table}[H]
    \centering
    \caption{Lorenz 96 system (Case 1): Performance of the models in the test period.}
    \label{tab:L96(case1)Error}
    \begin{tabular}{|c|c|c|c|}
        \hline
        &Forecast MSE &DA MSE & DA Neg-Log-Likelihood \\
        \hline
        Knowledge-based regression model &3.8283 & 13.7121 & 33.8079 \\
         \hline
        CGNSDE without DA loss &0.0410  & 0.0764 & 3.9665   \\
        \hline
        CGNSDE with the DA loss &0.0506 &0.0690 & 2.9490\\
        \hline
    \end{tabular}
\end{table}



\begin{figure}[H]
    \centering
    \includegraphics[width=0.7\textwidth]{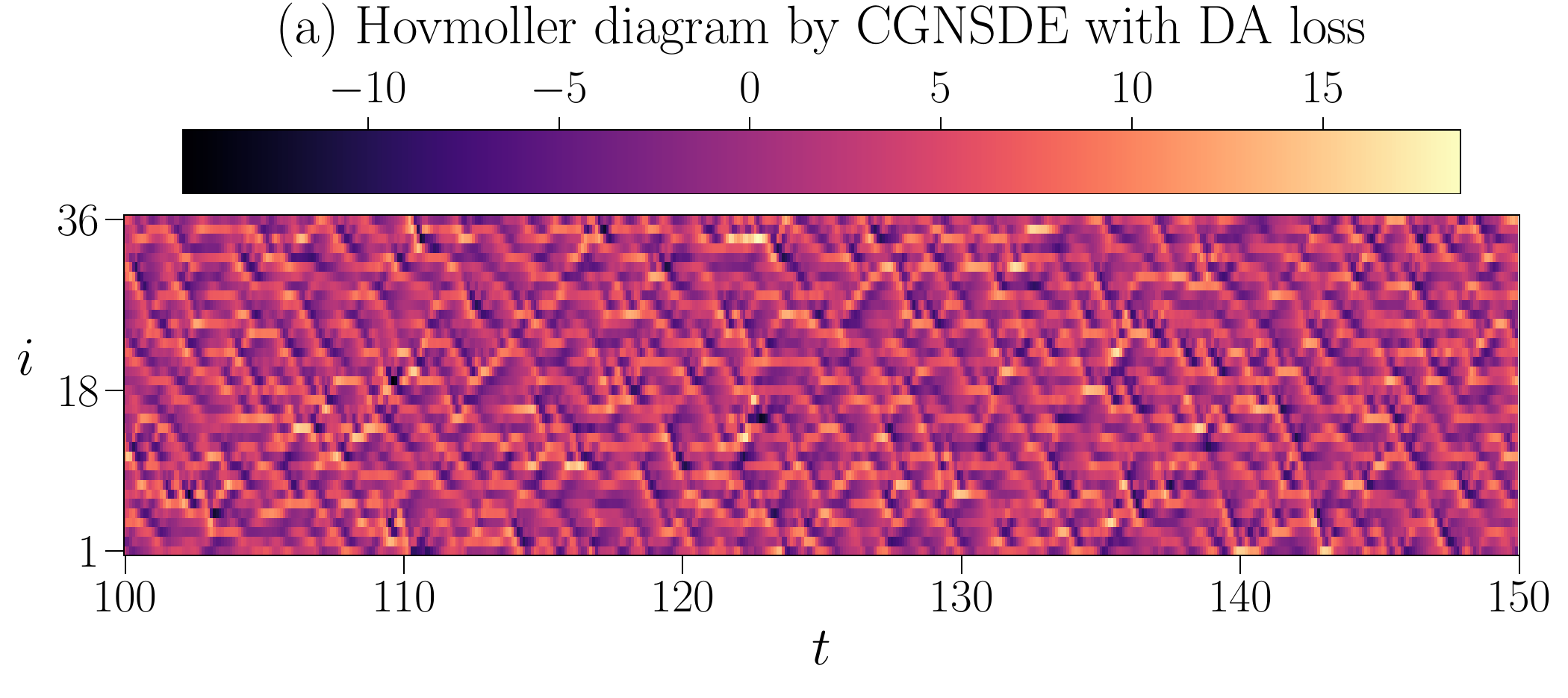}
    \includegraphics[width=\textwidth]{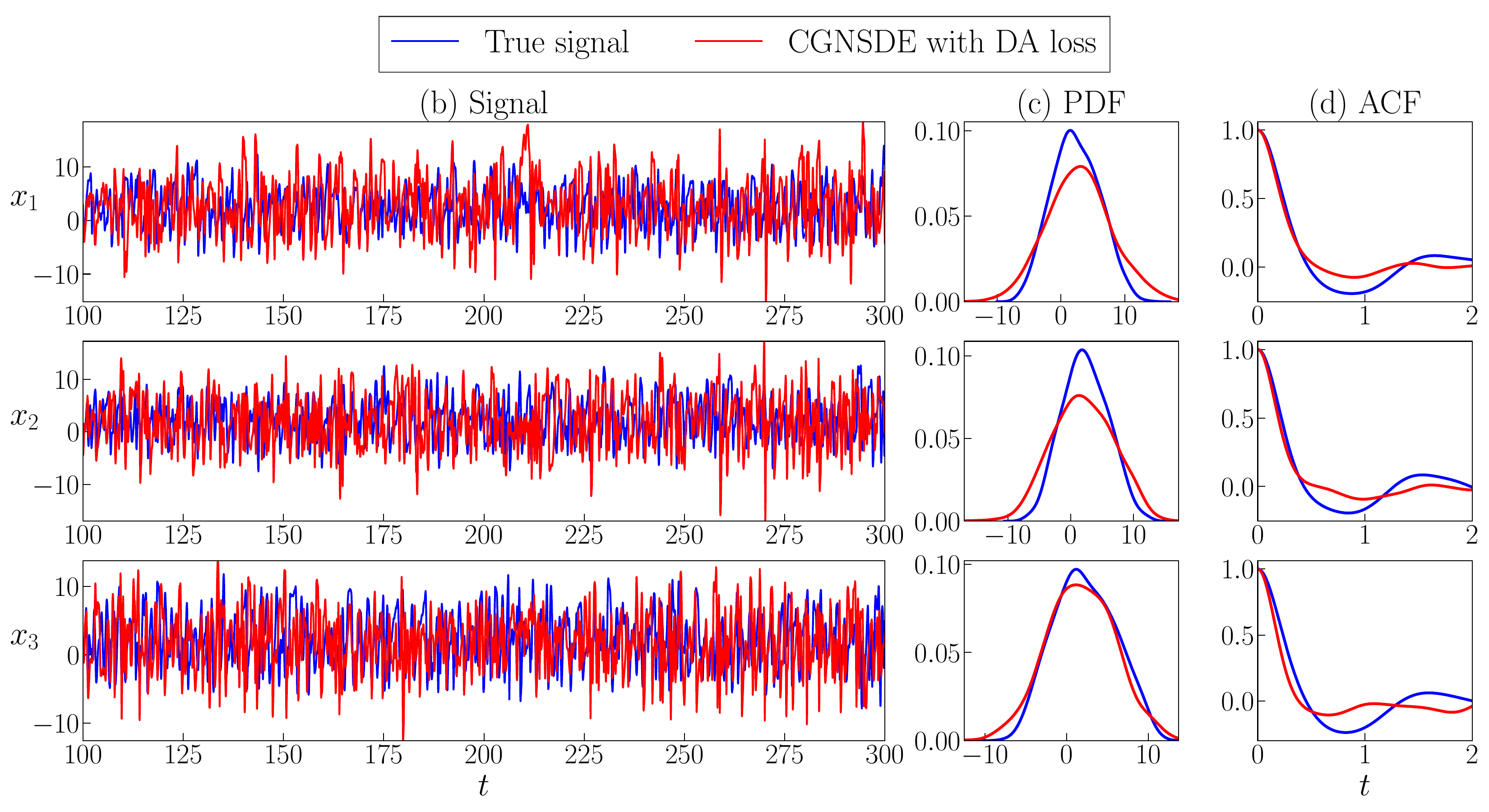}
    \caption{Long-term simulation results of CGNSDE with the DA loss for the Lorenz 96 system (Case 1). Panel (a): Hovmoller diagram of all states. Panel (b): time series of the first three states. Panel (c) and (d): probability density function (PDF) and auto-correlation function (ACF).}
    \label{fig:L96(case1)_LongSimu}
\end{figure}

Figure \ref{fig:L96(case1)_LongSimu} shows the long-term simulation of CGNSDE with the DA loss and compares the results with the true system. As shown in Panel (a), the CGNSDE model can produce stable long-term simulation, which captures the wave propagation and the chaotic pattern of the underlying dynamics. Since the CGNSDE model in \eqref{eq:L96(case1)_CGNSDE} has different types of expressions, one state of each kind is presented in Panel (b). The model successfully captures the overall patterns of the true time series, with a slight overestimate of the amplitude in $x_1$ and $x_2$. Panels (c) and (d) show the associated PDFs and the ACFs, confirming that the long-term statistics from the CGNSDE model are overall comparable to the truth, even though the forecast loss in the training period is based only on a much shorter time series data.

\subsubsection{Case 2: Non-CGNS that is spatially homogeneous}
\label{sssec:results_L96_2}

In this case, $\mathbf{u}_1 = [x_1, x_3, x_5, \cdots, x_{35}] \in \mathbb{R}^{18}$ are the observed variables and $\mathbf{u}_2 = [x_2, x_4, \cdots, x_{36}] \in \mathbb{R}^{18}$ are unobserved. By doing so, the true Lorenz 96 system $[\mathbf{u}_1, \mathbf{u}_2]^\mathtt{T}$ becomes a non-CGNS.

The causal inference method is utilized to identify significant knowledge-based components. The function libraries \eqref{eq:Library} are different for state variables $\mathbf{u}_1$ and $\mathbf{u}_2$ to guarantee the overall conditional Gaussian structures. The results of causation entropy are summarized in Table \ref{tab:L96(case2)_CEM}. The selected quadratic basis for $x_i\in\mathbf{u}_1$ are $[x_i, x_{i+1}, x_{i-2}x_{i-1}, x_{i-1}x_{i+2}, x_ix_{i+1}]$, while for $x_i\in\mathbf{u}_2$ are $[x_i, x_{i-2}x_{i-1}, x_{i-1}x_{i+1}]$.

\begin{table}[H]
\caption{Lorenz 96 system (Case 2): causation entropy from the candidate functions contributing to the dynamics. The significant values are highlighted in bold font.}
\label{tab:L96(case2)_CEM}
\centering
\begin{adjustbox}{max width=1.\textwidth,center}
\begin{tabular}{|c|c|c|c|c|c|c|c|c|c|}
\hline
\multirow{4}{*}{\rotatebox[origin=c]{90}{\parbox{2.0cm}{\centering $\dot{x}_i$\\ \footnotesize{$i\bmod 2 = 1$} }}} & $x_{i-2}$ & $x_{i-1}$ & $x_{i}$ & $x_{i+1}$ & $x_{i+2}$ & $x_{i-2}^2$ & $x_{i}^2$ & $x_{i+2}^2$ & $x_{i-2}x_{i-1}$  \\ \cline{2-10}
& 0.000  & 0.007 & \textbf{0.013} & \textbf{0.025} & 0.001& 0.006& 0.003& 0.001& \textbf{0.247} \\ \cline{2-10}
& $x_{i-2}x_{i}$ & $x_{i-2}x_{i+1}$ & $x_{i-2}x_{i+2}$ & $x_{i-1}x_{i}$ & $x_{i-1}x_{i+2}$ & $x_ix_{i+1}$ & $x_ix_{i+2}$ & $x_{i+1}x_{i+2}$ & - \\ \cline{2-10}
& 0.003 & 0.007 & 0.003 & 0.008 & \textbf{0.014} & \textbf{0.015} & 0.004 & 0.001 & -  \\ \hline
\hline
\multirow{4}{*}{\rotatebox[origin=c]{90}{\parbox{2cm}{\centering $\dot{x}_i$\\ \footnotesize{$i\bmod 2 = 0$} }}} & $x_{i-2}$ & $x_{i-1}$ & $x_{i}$ & $x_{i+1}$ & $x_{i+2}$ & $x_{i-1}^2$ & $x_{i+1}^2$ & $x_{i-2}x_{i-1}$ & $x_{i-2}x_{i+1}$  \\ \cline{2-10}
& 0.000 & 0.000 & \textbf{0.085} & 0.000 & 0.000 & 0.001 & 0.000 & \textbf{0.820} & 0.000  \\ \cline{2-10}
& $x_{i-1}x_i$ & $x_{i-1}x_{i+1}$ & $x_{i-1}x_{i+2}$ & $x_ix_{i+1}$ & $x_{i+1}x_{i+2}$ & - & - & - & -  \\ \cline{2-10}
 &0.000  & \textbf{0.726} & 0.000  &  0.000 &  0.000 & - & - &- &-   \\ \hline
\end{tabular}
\end{adjustbox}
\end{table}

With the selected candidate functions, the knowledge-based regression model (KRM) is given by:
\begin{equation}
    \label{eq:L96(case2)_KRM}
    \frac{\diff x_i}{\diff t} =
    \begin{cases}
        \text{ if } i\bmod 2 = 1: \\
        \quad \mathbf{KRM}_1 = a_0 + a_1x_i + a_2x_{i+1} + a_3x_{i-2}x_{i-1} + a_4x_{i-1}x_{i+2}+a_5x_ix_{i+1}+\sigma_i\dot{W}_i\\
        \text{ if } i\bmod 2 = 0: \\
        \quad \mathbf{KRM}_2 = b_0 + b_1x_i + b_2x_{i-2}x_{i-1} + b_3x_{i-1}x_{i+1}+\sigma_i\dot{W}_i.
    \end{cases} \\
\end{equation}
After introducing the neural network components, the CGNSDE can be written in the form:
\begin{equation}
    \label{eq:L96(case2)_CGNSDE}
    \frac{\diff x_i}{\diff t} =
    \begin{cases}
        \text{ if } i\bmod 2 = 1: \\
        \quad \begin{aligned}
            \mathbf{KRM}_1 + &\mathbf{NN}^{(1)}_1(x_{i-2}, x_i, x_{i+2};\theta_1) +\\
            &\mathbf{NN}^{(1)}_2(x_{i-2}, x_i, x_{i+2};\theta_1)x_{i-1}+ \\
            &\mathbf{NN}^{(1)}_3(x_{i-2}, x_i, x_{i+2};\theta_1)x_{i+1},
            \end{aligned}\\
        \text{ if } i\bmod 2 = 0: \\
        \quad \begin{aligned}
        \mathbf{KRM}_2 + &\mathbf{NN}^{(2)}_1(x_{i-1},x_{i+1};\theta_2) + \\
        &\mathbf{NN}^{(2)}_2(x_{i-1},x_{i+1};\theta_2)x_{i-2} + \\
        &\mathbf{NN}^{(2)}_3(x_{i-1},x_{i+1};\theta_2)x_i + \\
        &\mathbf{NN}^{(2)}_4(x_{i-1},x_{i+1};\theta_2)x_{i+2},
        \end{aligned}
    \end{cases}
\end{equation}
where again the localization is imposed, i.e., $\diff x_i/\diff t$ only depends on nearby states $(x_i,x_{\pm 1},x_{\pm 2})$. The $\mathbf{NN}^{(i)}_i$ in \eqref{eq:L96(case2)_CGNSDE} stands for the $j$-th output of $i$-th neural network. The two neural networks have 5 layers with 438 parameters and 3 layers with 70 parameters, respectively.

With the same training and testing settings listed in Case 1, the test results of three models are summarized in Table \ref{tab:L96(case2)_Error}. The CGNSDE without DA loss can outperform the knowledge-based regression model in both state forecast and DA. After retraining with the DA loss, the CGNSDE can further reduce the DA MSE and the negative log-likelihood at the expense of a slight increment in the forecast MSE. The DA MSE and DA negative log-likelihood of the reference solution from the EnKBF with the true system are 0.0771 and 13.9382, respectively. 

\begin{table}[H]
    \centering
    \caption{Lorenz 96 system (Case 2): Performance of the models in the test period.}
    \label{tab:L96(case2)_Error}
    \begin{tabular}{|c|c|c|c|}
        \hline
        &Forecast MSE &DA MSE & DA Neg-Log-Likelihood \\
        \hline
        Knowledge-based regression model & 1.1137 & 1.3592 & 58.9987 \\
         \hline
        CGNSDE without DA loss & 0.7952& 0.9710 & 54.5143 \\
        \hline
        CGNSDE with the DA loss & 0.9482 & 0.5763 & 30.0057\\
        \hline
    \end{tabular}
\end{table}

\begin{figure}[H]
    \centering
    \includegraphics[width=\textwidth]{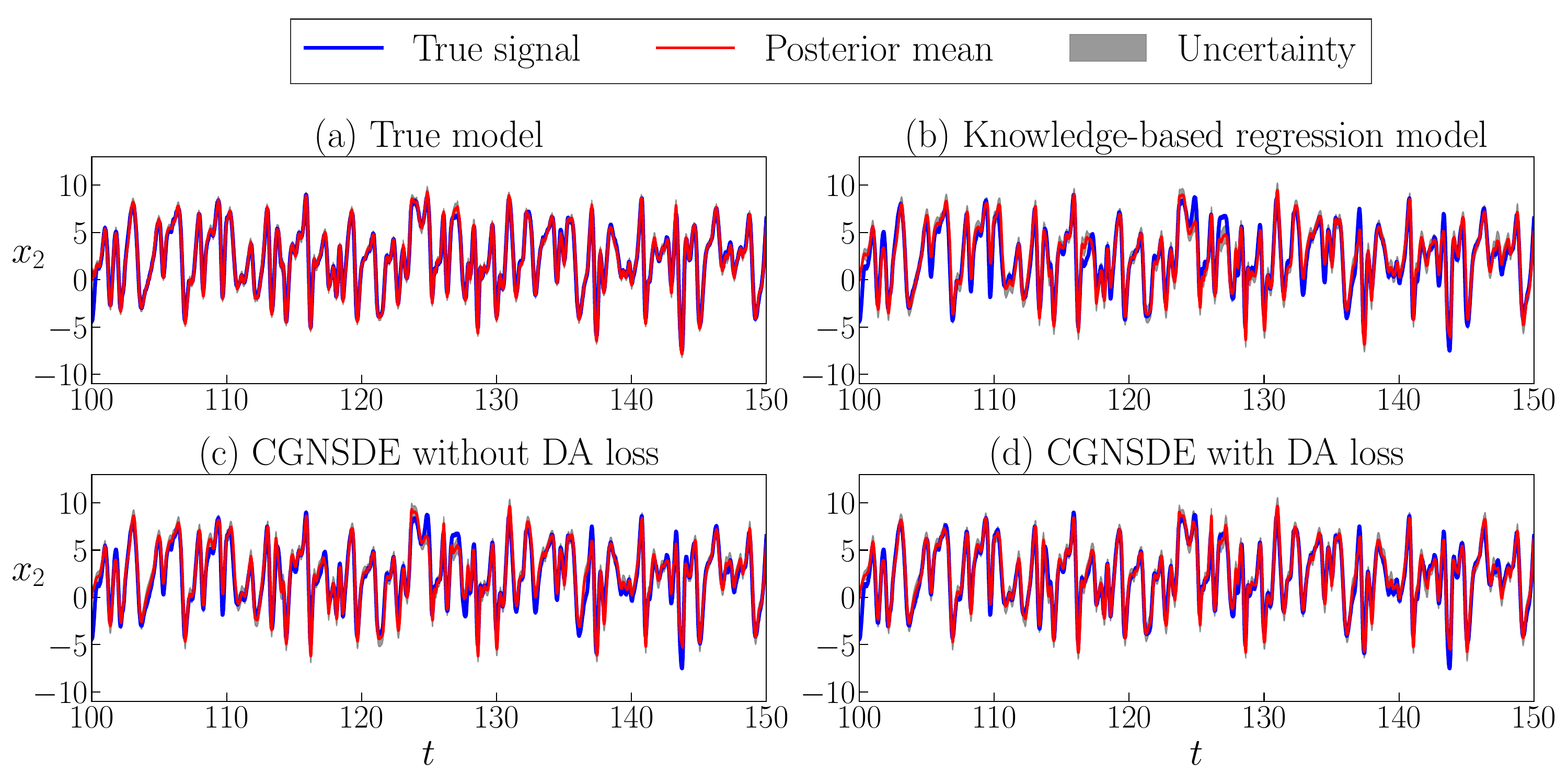}
    \caption{DA results of the unobserved state $x_2$ in the Lorenz 96 system (Case 2) for the true system (using the EnKBF) and for the other three models. The uncertainties are indicated by the grey colored regions, which correspond to two standard deviation from the posterior mean.}
    \label{fig:L96(case2)_DA}
\end{figure}

Figure \ref{fig:L96(case2)_DA} compares the DA results of the unobserved state $x_2$ from the regression model and the two CGNSDEs to the true system using the EnKBF. Although the CGNSDE with the DA loss is still slightly worse than the perfect system, it is more skillful than the regression model and the CGNSDE without the DA loss. It is also worth noting that, despite some errors, the regression model gives an acceptable result. This demonstrates that the explainable knowledge-based components determined by the causation entropy indeed play an essential role in modeling the system and DA.

\begin{figure}[H]
    \centering
    \includegraphics[width=0.7\textwidth]{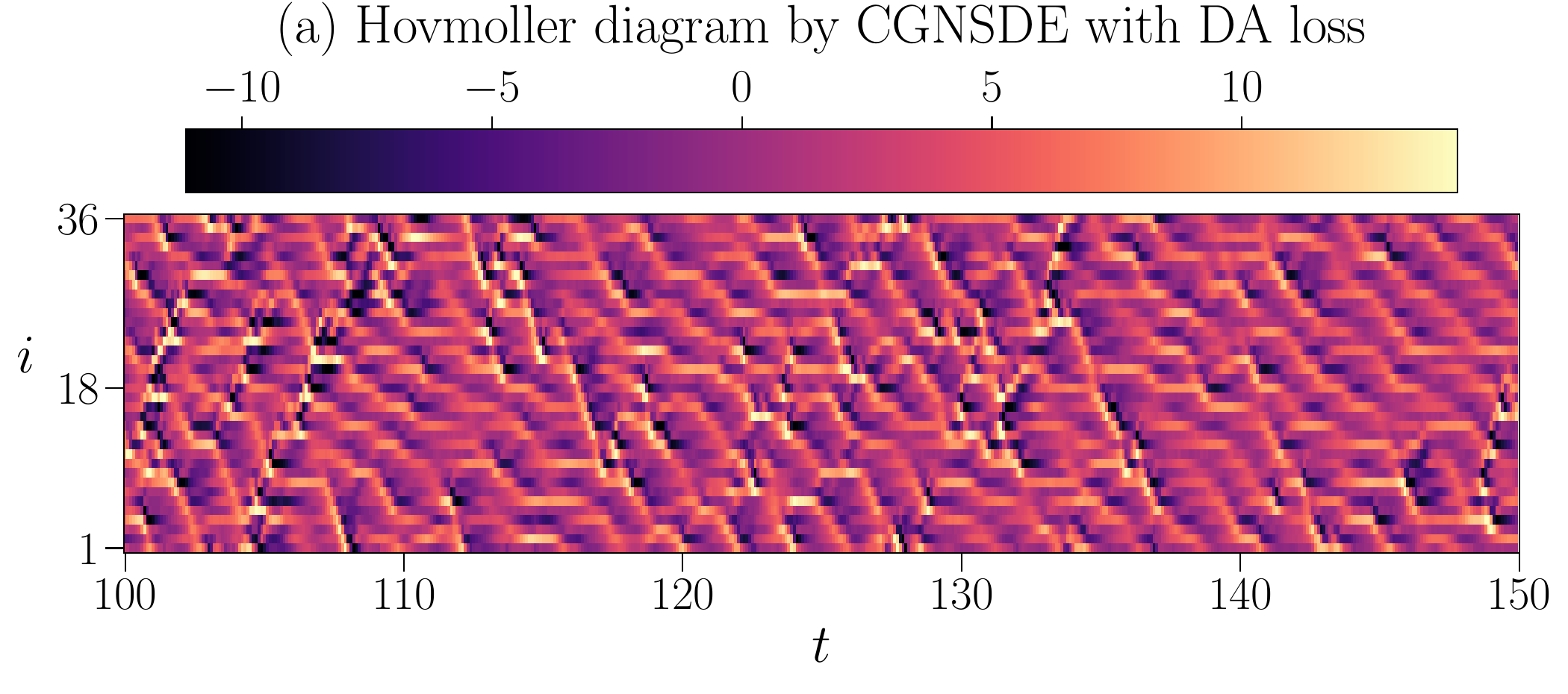}
    \includegraphics[width=\textwidth]{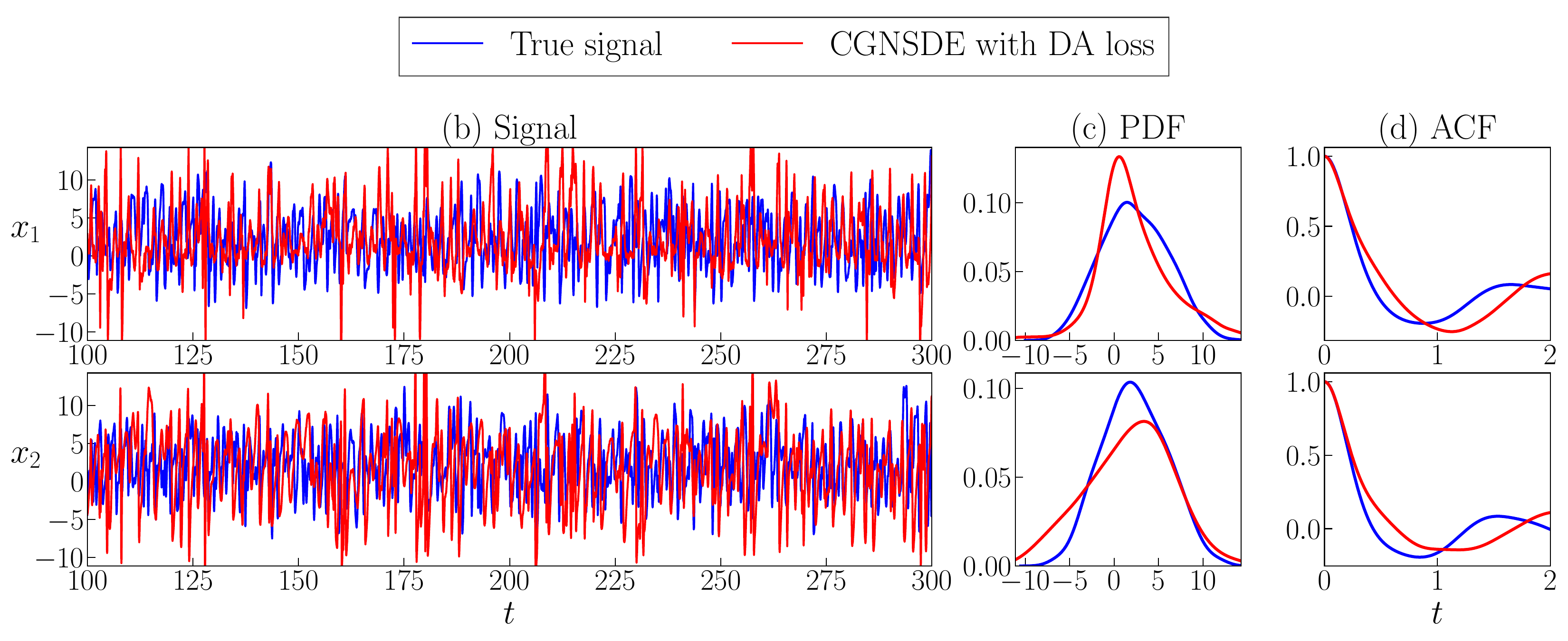}
    \caption{Long-term simulation results of the CGNSDE with the DA loss for the Lorenz 96 system (Case 2). Panel (a): Hovmoller diagram of all states. Panel (b): time series of the first three states. Panel (c) and (d): The PDFs and the ACFs.}
    \label{fig:L96(case2)_LongSimu}
\end{figure}

Figure \ref{fig:L96(case2)_LongSimu} shows the long-term simulation results of CGNSDE with the DA loss. Since the CGNSDE has very different dynamics from the truth of such a highly chaotic system and is trained based only on short-term forecast loss, it is not expected that the CGNSDE will fully capture the dynamical features of the truth. Nevertheless, Panel (a) illustrates that the simulation from the CGNSDE preserves the chaotic features and irregular wave propagations. The time series, as shown in Panel (b), validates the chaotic nature of the simulation from the CGNSDE. Although the statistics shown in Panels (c) and (d) contain some error, the overall variability and temporal dependence are reproduced by the CGNSDE.

\subsubsection{Case 3: Non-CGNS that is spatially inhomogeneous}
\label{sssec:results_L96_3}

The last case involves a spatially inhomogeneous case in the true system, which does not satisfy the CGNS. The two neural networks have 5 layers with 543 parameters and 3 layers with 119 parameters, respectively.

Table \ref{tab:L96Inhomo_Error} summarizes the model performance in the test period. The two CGNSDE models can improve the performance of state forecast and DA compared with the knowledge-based regression model. The CGNSDE with the DA loss in training has the lowest DA error for the test data, with some trade-offs in forecasting accuracy compared to the CGNSDE without DA loss. The DA MSE and DA negative log-likelihood of applying EnKBF to this true inhomogeneous system are 0.0672 and 8.8427, respectively. 

\begin{table}[H]
    \centering
    \caption{Lorenz 96 system (Case 3): Performance of the models in the test period.}
    \label{tab:L96Inhomo_Error}
    \begin{tabular}{|c|c|c|c|}
        \hline
        &Forecast MSE & DA MSE & DA Neg-Log-Likelihood \\
        \hline
        Knowledge-based regression model &1.1625 & 1.2634& 58.0666 \\
         \hline
        CGNSDE without DA loss & 0.2984 & 0.4903 & 38.1015  \\
        \hline
        CGNSDE with the DA loss & 0.3752  & 0.3794 &26.2498 \\
        \hline
    \end{tabular}
\end{table}

\begin{figure}[H]
    \centering
    \includegraphics[width=0.7\textwidth]{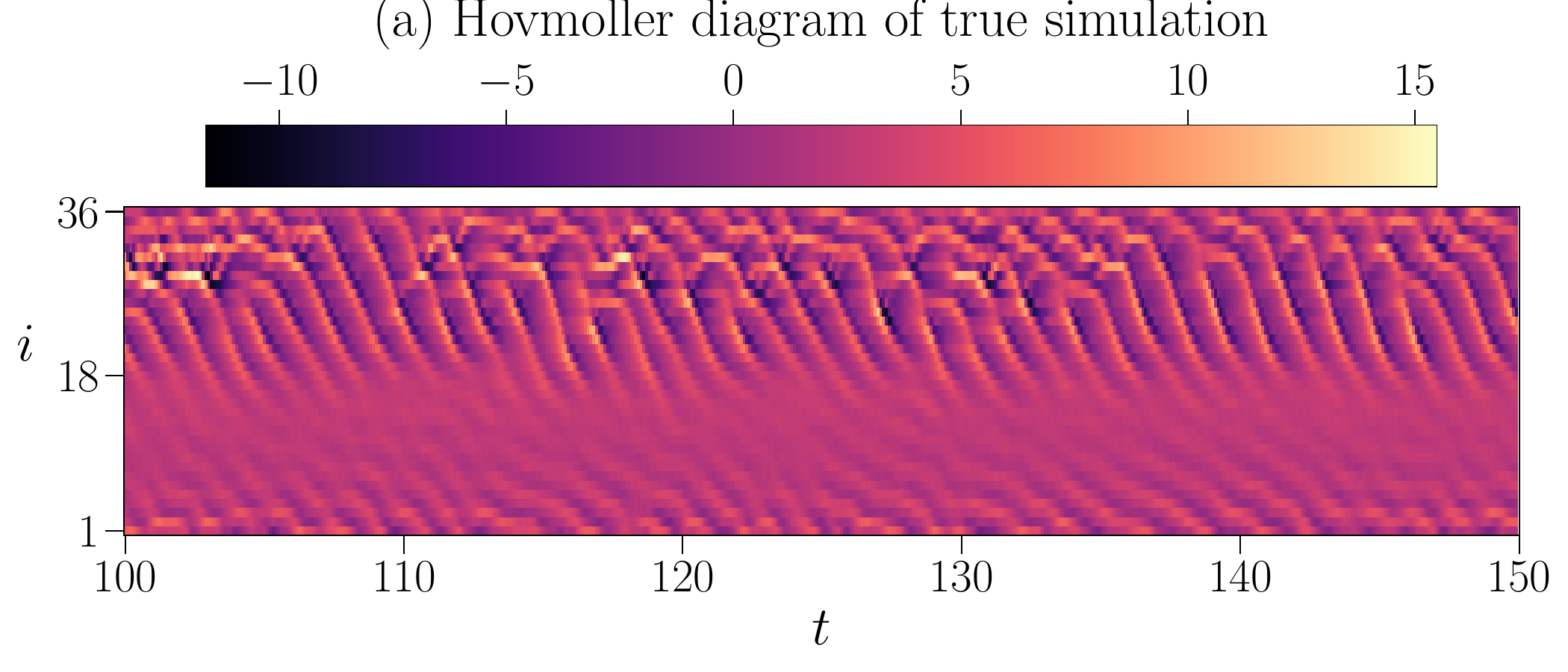}

    \includegraphics[width=\textwidth]{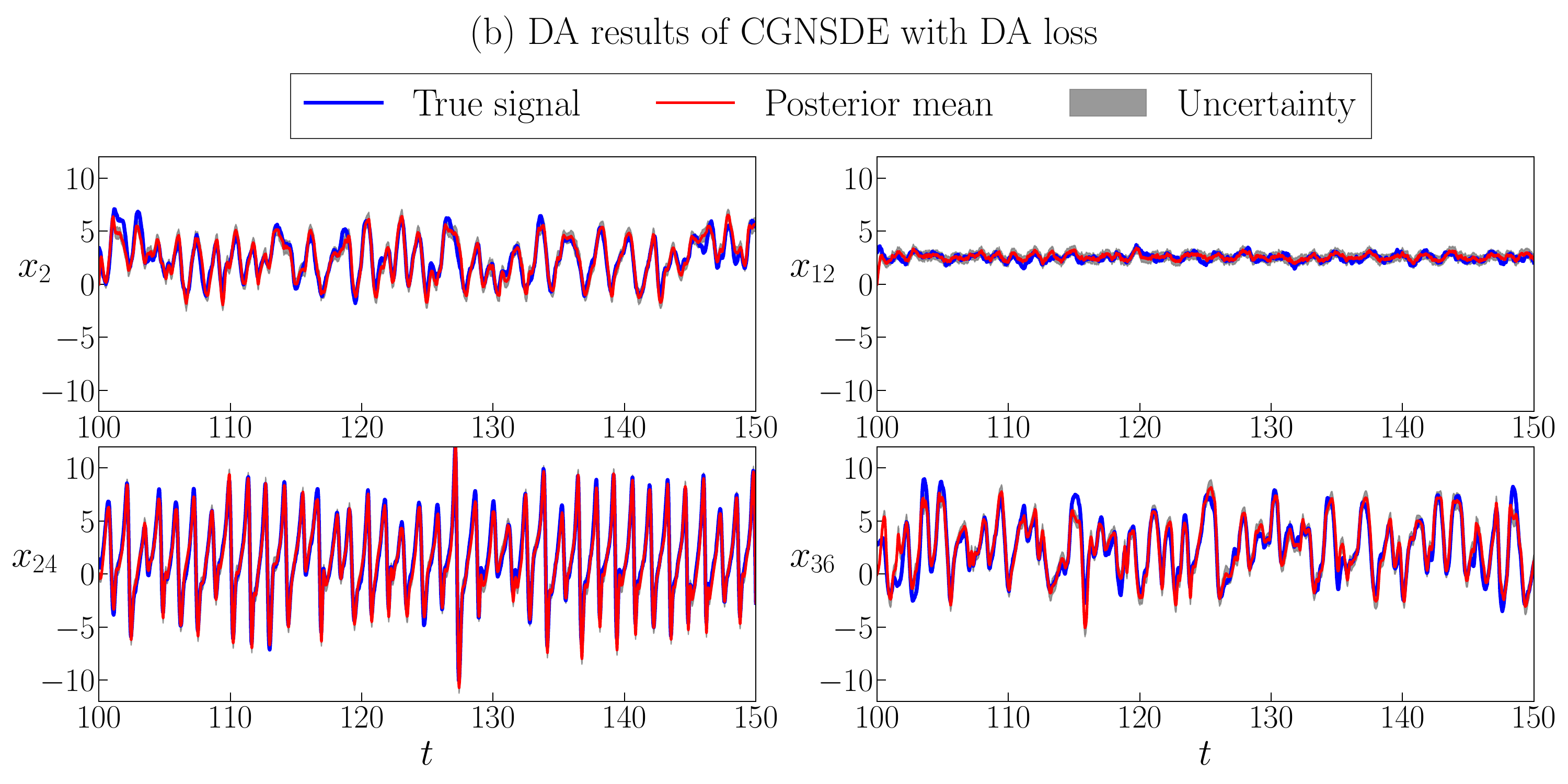}
    \caption{Model simulation and the DA results of the inhomogeneous Lorenz 96 system (Case 3). Panel (a): Hovmoller diagram of all states in the true system. Panel (b): DA results of the unobserved states $x_2$, $x_{12}$, $x_{24}$, and $x_{36}$ using the CGNSDE with the DA loss. The associated uncertainties are indicated by the grey colored regions, which correspond to two standard deviation from the posterior mean.
    }

    \label{fig:L96Inhomo}
\end{figure}

Panel (a) of Figure \ref{fig:L96Inhomo} displays the inhomogeneous spatiotemporal pattern of the true system. Panel (b) shows DA results using the CGNSDE trained with the DA loss for several states ($x_2$, $x_{12}$, $x_{24}$ and $x_{36}$), whose indices are distributed across the location range of the system. The DA results are overall accurate. Note that half of the grid points with the odd/even index numbers still share the same neural networks as in \eqref{eq:L96(case2)_CGNSDE}. Such a computationally efficient strategy remains working well even for this inhomogeneous case.

\section{Discussions and Conclusion}\label{Sec:Conclusion}
\subsection{Conclusion}
In this paper, a new knowledge-based and machine-learning hybrid modeling approach, the CGNSDE, is developed. It aims to facilitate modeling complex dynamical systems and implementing the associated DA. In contrast to the standard neural network predictive models, the CGNSDE is designed to effectively tackle both forward prediction tasks and inverse state estimation problems. The unique features of the CGNSDE are summarized as follows:
\begin{enumerate}
  \item The CGNSDE exploits a computationally robust causal inference method to develop a parsimonious model with knowledge-based components that correspond to interpretable physics.
  \item Neural networks are supplemented to such knowledge-based components in a systematic way that allows closed analytic formulae for efficient DA.
  \item The analytic formulae are used in an additional computationally affordable loss to train the neural networks that explicitly improve the accuracy of the CGNSDE for DA.
  \item The DA loss function also advances the neural networks to enhance the performance of identifying the causal relationship of the underlying system, further strengthening the modeling skills.
  \item The CGNSDE is more capable of estimating extreme events and quantifying the associated uncertainty.
  \item Crucial physical properties in many complex systems, such as the translate-invariant local dependence of state variables, can significantly simplify the neural network structures and facilitate the CGNSDE to be applied to high-dimensional systems.
\end{enumerate}
Utilizing a hierarchy of numerical experiments, it is shown that the CGNSDE is skillful in modeling and estimating the state of chaotic systems with intermittency and strong non-Gaussian features. In particular, the additional neural network components allow the CGNSDE to significantly outperform the standard regression models. It is also demonstrated that the DA loss in the CGNSDE plays a vital role in enhancing the skill of state estimation, which is not always highlighted in state-of-the-art neural network predictive models.

There are quite a few extensions of the current CGNSDE framework, which provides the potential for the approach to be further improved and applied to a wide range of challenging problems.

\subsection{Other forms of the DA loss function}
One of the crucial features of the CGNSDE is the additional DA loss. In this paper, the loss is designed in the simplest possible way by minimizing the error in the posterior mean time series related to the truth. Since the posterior mean depends on the posterior covariance in \eqref{eq:CGNS_Filter}, the information in the entire posterior distribution is still utilized in such a setup. Nevertheless, the closed analytic formulae for both the mean and covariance can be exploited more delicately. One natural idea is to use the likelihood function as the DA loss \eqref{eq:DA_NLL}, as was utilized as the validation criterion in the numerical experiment section. The likelihood can be regarded as a weighted MSE, where the square root of the time-varying covariance serves as the weight at each time instant. One obvious advantage of using likelihood loss is that uncertainty will be characterized more rigorously, especially when quantifying uncertainty and intermittency.

\subsection{CGNSDE with recurrent neural network}
The neural networks in all the numerical experiments of this study are the standard feed-forward networks. Such a simple setup allows us to verify the crucial role of each unique feature of the CGNSDE in improving modeling and DA skills. In practice, many other types of neural networks can be incorporated into the CGNSDE. Particularly, including the memory in the signals is helpful for prediction and state estimation. Since the DA formulae in \eqref{eq:CGNS_Filter} can depend on the entire history of the observed variables $\mathbf{u}_1(s\leq t)$, it justifies the incorporation of the memory effects into the neural networks in \eqref{eq:CGNSDE_Details}. One of the most natural ways in this direction is to adopt recurrent neural networks \citep{medsker2001recurrent, hochreiter1997long} in the CGNSDE. Other advanced techniques, such as the transformer \citep{vaswani2017attention, bommasani2021opportunities}, can also be included in the CGNSDE.

\subsection{Discrete-in-time extension}
Another natural extension of the current framework is considering the situation with discrete-in-time observations, which appear in many geophysical applications. This is an analog to the continuous-in-time DA framework in this paper \eqref{eq:CGNS_Filter}. Closed analytic formulae still exist when the observations arrive at discrete instants \citep{liptser2013statistics}.

\section*{Acknowledgments}
The research of N.C. was partially funded by the Office of Naval Research (ONR) N00014-24-1-2244. The research of C.C. and J.W. was funded by the University of Wisconsin-Madison, Office of the Vice Chancellor for Research and Graduate Education with funding from the Wisconsin Alumni Research Foundation.

\section*{Data Availability}
The data that support the findings of this study are available from the corresponding author upon reasonable request. The codes and examples that support the findings of this study are available in the link: \url{https://github.com/ChuanqiChenCC/CGNSDE}.

\bibliographystyle{unsrt}
\bibliography{references}

\end{document}